\documentclass[journal]{IEEEtran}

\usepackage{times}
\usepackage{epsfig}
\usepackage{graphicx}
\usepackage{amsmath}
\usepackage{amssymb}
\usepackage{multirow}
\usepackage{cite}
\usepackage{array}
\usepackage{xcolor}
\usepackage{float}
\usepackage{url}
\usepackage{soul}
\usepackage{bbm}
\usepackage{mathtools}

\ifCLASSINFOpdf
\else
\fi

\newfloat{figtab}{t}{t}
\makeatletter
\newcommand\figcaption{\def\@captype{figure}\caption}
\newcommand\tabcaption{\def\@captype{table}\caption}
\makeatother

\begin{document}

\indent

© 2023 IEEE. Personal use of this material is permitted. Permission from IEEE must be obtained for all other uses, in any current or future media, including reprinting/republishing this material for advertising or promotional purposes, creating new collective works, for resale or redistribution to servers or lists, or reuse of any copyrighted component of this work in other works. Digital Object Identifier 10.1109/TIP.2023.3298537

%
\title{SVCNet: Scribble-based Video Colorization Network with Temporal Aggregation}
%
%
%

\author{Yuzhi~Zhao,~\IEEEmembership{Graduate~Student~Member,~IEEE,}
        Lai-Man~Po,~\IEEEmembership{Senior~Member,~IEEE,}
        Kangcheng~Liu,~\IEEEmembership{Member,~IEEE,} Xuehui~Wang,~\IEEEmembership{Student~Member,~IEEE}, Wing-Yin~Yu,~\IEEEmembership{Student~Member,~IEEE}, Pengfei Xian,~\IEEEmembership{Student~Member,~IEEE}, Yujia Zhang, Mengyang Liu 
\thanks{Manuscript received November 14, 2022, revised March 21, 2023, accepted July 20, 2023. \textit{(Corresponding author: Yuzhi Zhao.)}}
\thanks{Y. Zhao, L.-M. Po, W.-Y. Yu, and P. Xian are with the Department of Electronic Engineering, City University of Hong Kong, Hong Kong, China (e-mail: yzzhao2-c@my.cityu.edu.hk; eelmpo@cityu.edu.hk; wingyinyu8-c@my.cityu.edu.hk; xian.pf@my.cityu.edu.hk).}
\thanks{K. Liu is with the School of Computer Science and Engineering, Nanyang Technological University, Singapore. (email: kangcheng.liu@ntu.edu.sg).}
\thanks{X. Wang is with the Artificial Intelligence Institute, Shanghai Jiao Tong University, Shanghai, China (e-mail: wangxuehui@sjtu.edu.cn).}
\thanks{Y. Zhang and M. Liu are with Tencent Video, Tencent Holdings Ltd, Shenzhen, China (e-mail: yujiazhang@tencent.com; myleonliu@tencent.com).}
}

%
%

\markboth{IEEE Transactions on Image Processing}%
{Shell \MakeLowercase{\textit{Zhao \emph{et al.}}}: SVCNet: Scribble-based Video Colorization Network with Temporal Aggregation}
%



\maketitle

\begin{abstract}

In this paper, we propose a scribble-based video colorization network with temporal aggregation called SVCNet. It can colorize monochrome videos based on different user-given color scribbles. It addresses three common issues in the scribble-based video colorization area: colorization vividness, temporal consistency, and color bleeding. To improve the colorization quality and strengthen the temporal consistency, we adopt two sequential sub-networks in SVCNet for precise colorization and temporal smoothing, respectively. The first stage includes a pyramid feature encoder to incorporate color scribbles with a grayscale frame, and a semantic feature encoder to extract semantics. The second stage finetunes the output from the first stage by aggregating the information of neighboring colorized frames (as short-range connections) and the first colorized frame (as a long-range connection). To alleviate the color bleeding artifacts, we learn video colorization and segmentation simultaneously. Furthermore, we set the majority of operations on a fixed small image resolution and use a Super-resolution Module at the tail of SVCNet to recover original sizes. It allows the SVCNet to fit different image resolutions at the inference. Finally, we evaluate the proposed SVCNet on DAVIS and Videvo benchmarks. The experimental results demonstrate that SVCNet produces both higher-quality and more temporally consistent videos than other well-known video colorization approaches. The codes and models can be found at \url{https://github.com/zhaoyuzhi/SVCNet}.

\end{abstract}


%
\IEEEpeerreviewmaketitle

\section{Introduction}

Video colorization is the process of attaching plausible colors to monochrome videos. Restricted by imaging technology, many old films are preserved in black-and-white format. It is highly desirable for people to watch colorful videos. Recently, deep neural networks have achieved great improvements in both video restoration and colorization areas. Therefore, recovering realistic and colorful videos with deep neural networks becomes plausible.

The main difficulties for video colorization are \emph{colorization vividness} and \emph{temporal consistency} of sequential frames. Besides, \emph{color bleeding} (i.e., the spreading of colors beyond the object boundary) is another challenge. There are many solutions for existing methods to these problems, which typically fall into one of these four categories:

1) Image colorization and temporal smoothing \cite{bonneel2015blind, lai2018, zhou2020temporal, lei2020blind};

2) Image colorization and color propagation \cite{jampani2017video, meyer2018deep, endo2020practical, wu2020memory};

3) Fully-automatic video colorization \cite{lei2019fully, kouzouglidis2019automatic, zhao2022vcgan};

4) Exemplar-based video colorization \cite{zhang2019deep, iizuka2019deepremaster, wan2022bringing}.

The first three categories are not based on additional guidance such as exemplar images and color scribbles. To learn the grayscale to color mapping, they normally adopt large training sets like ImageNet \cite{russakovsky2015imagenet} to learn rich data priors. The differences between the three categories are obvious. Since category 1) relies on pre-trained image colorization methods and only finetunes the single image colorization results, video continuity cannot be ensured. Category 2) is similar to category 1) but only uses the first colorized frame. It depends on the long-range connection too much and easily ignores the characteristics of every single frame. Category 3) performs better than 1) and 2) since it jointly learns colorization and temporal smoothing. However, they are difficult to predict realistic color embeddings since the grayscale format losses too much information compared with the color format (e.g., \emph{RGB}, \emph{YUV}, and \emph{CIE Lab}). Furthermore, they may sacrifice colorfulness due to temporal constraints. To improve colorization quality, category 4) induces an exemplar image to guide the video colorization. Though it produces more colorful videos than other methods, it requires a relatively accurate exemplar image similar to the color version of the monochrome input.

\begin{figure*}[t]
\begin{center}
\centering
\includegraphics[width=0.95\linewidth]{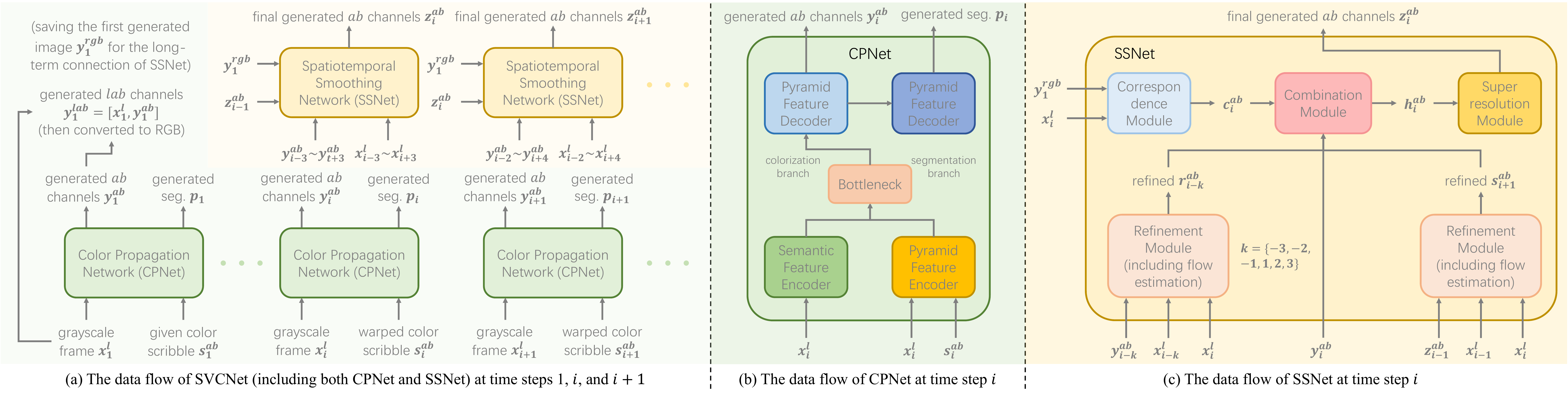}
\caption{Illustration of the data flows of (a) SVCNet (including CPNet and SSNet), (b) CPNet, and (c) SSNet. Users only need to provide color scribbles for the first frame $s_1^{ab}$. The following color scribbles $s_2^{ab}$-$s_t^{ab}$ are obtained by warping $s_1^{ab}$ using the forward optical flows.}
\label{pipeline}
\end{center}
\end{figure*}

To achieve higher video colorization quality than existing methods and minimize color bleeding artifacts, we propose the first scribble-based video colorization framework called SVCNet, where the data flow is illustrated in Figure \ref{pipeline}. Compared with previous solutions, there are four improvements:

1) Improved colorization vividness: SVCNet includes two stages for precise colorization (CPNet) and temporal smoothing (SSNet), respectively. The CPNet is a multi-input-multi-output architecture that achieves better colorization quality;

2) Strengthened temporal consistency: SVCNet includes both bidirectional projection and long-range connection, which effectively aggregate the temporal information of the video;

3) Reduced color bleeding artifacts: SVCNet learns an additional auxiliary segmentation task to minimize color bleeding artifacts besides performing video colorization;

4) Reduced manual work: SVCNet only needs sparse color scribbles instead of an accurate exemplar image.

SVCNet includes two sub-networks: color propagation sub-network (CPNet) and spatiotemporal smoothing sub-network (SSNet), respectively. Its data flow is shown in Figure \ref{pipeline}. Firstly, CPNet is a multi-input-multi-output architecture, as shown in Figure \ref{pipeline} (b). It includes two encoders to extract semantics from the grayscale input and combine the information of both inputs, which performs precise scribble-color propagation and ensures good colorization quality when there are even no color scribbles. In addition, it has two decoder branches, where the colorization branch outputs color embeddings and the other outputs the corresponding segmentation map. The segmentation branch assists the colorization branch through backward propagation at the training. It helps the CPNet reduce the color bleeding artifacts since it pushes the network to separate clear boundaries. Secondly, SSNet refines every colorized frame of the CPNet, as shown in Figure \ref{pipeline} (c). On one hand, it performs the bidirectional projection based on previous, current, and leading frames by a Refinement Module, which serve as short-range connections. On the other hand, it extracts and aligns the first colorized frame as the long-range connection by a Correspondence Module. After that, a Combination Module aggregates all the information. Therefore, SSNet colorizes videos with satisfactory temporal consistency. We notice color embeddings (i.e., \emph{ab} channels in the \emph{CIE Lab} color space) are much sparser than edges. Based on this and also inspired by \cite{liu2010sift, chen2016bilateral, guadarrama2017pixcolor, zhao2022d2hnet}, we set all the aforementioned operations on a small and fixed image resolution and use a Super-resolution Module at the end of the SSNet to recover the original image resolution. Therefore, SVCNet fits different image resolutions at the inference. Finally, compared with exemplar-based video colorization, the proposed framework only needs sparse color scribbles input. Users do not need to select a proper exemplar image while only need to define the desired colors in some specific pixels.

We train and evaluate the SVCNet on both DAVIS \cite{perazzi2016benchmark} and Videvo \cite{Videvo} datasets. Extensive experiments demonstrate that SVCNet performs better than state-of-the-art video colorization methods. We also show that the CPNet achieves state-of-the-art scribble-based image colorization performance with fewer color bleeding artifacts. The main contributions of this paper are as follows:

1) We propose the first scribble-based video colorization framework called SVCNet, which includes two stages for color propagation and spatiotemporal smoothing, respectively. We set the most operations on a small and fixed image resolution to reduce the computational costs for producing videos with different large resolutions;
    
2) We propose a temporal aggregation method for video colorization including both short- and long-range connections;

3) We adopt a segmentation loss to address color bleeding artifacts in the video colorization area. Also, we generate saliency maps as pseudo-binary segmentation maps when there are no labeled segmentation maps in the datasets.

\begin{figure*}[t]
\begin{center}
\centering
\includegraphics[width=0.95\linewidth]{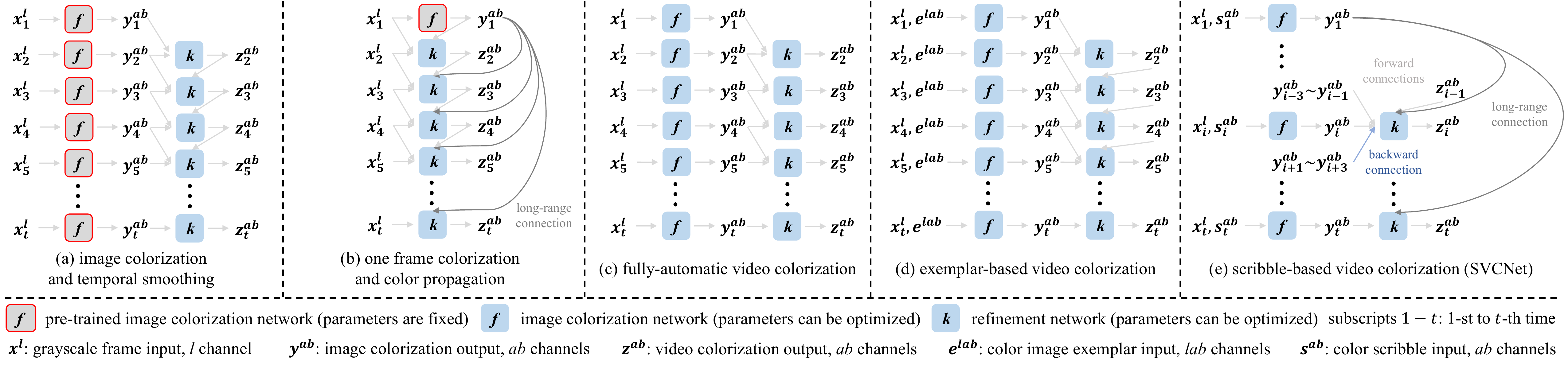}
\caption{Illustration of data flows of video colorization methods: (a) image colorization and temporal smoothing \cite{bonneel2015blind, lai2018, zhou2020temporal, lei2020blind}; (b) one frame colorization and color propagation \cite{jampani2017video, meyer2018deep, endo2020practical, wu2020memory}; (c) fully-automatic video colorization \cite{lei2019fully, kouzouglidis2019automatic, zhao2022vcgan}; (d) exemplar-based colorization \cite{zhang2019deep, iizuka2019deepremaster, wan2022bringing}; (e) scribble-based colorization (SVCNet), where only the first color scribble $s^{ab}_1$ is given by the user. The following color scribbles $s^{ab}_2$, $s^{ab}_3$, ... , $s^{ab}_t$ are obtained by warping the first color scribble $s^{ab}_1$ using forward optical flows $O_{1 \rightarrow 2}$, $O_{2 \rightarrow 3}$, ... , $O_{t-1 \rightarrow t}$.}
\label{compare_arch_video_colorization}
\end{center}
\end{figure*}

\section{Related work}

\noindent \textbf{Image Colorization.} Image colorization learns to reconstruct color embeddings from corresponding grayscale images. It can be categorized into reference-based colorization (e.g., scribble-based colorization \cite{levin2004colorization, huang2005adaptive, yatziv2006fast, xu2009efficient, chen2012manifold, xu2013sparse, sheng2013video, paul2016spatiotemporal, zhang2017real, zhang2018two, ci2018user, zhao2021legacy}, exemplar-based colorization \cite{reinhard2001color, welsh2002transferring, ironi2005colorization, tai2005local, charpiat2008automatic, chia2011semantic, bugeau2013variational, jampani2017video, meyer2018deep, he2018deep, sun2019adversarial, iizuka2019deepremaster, zhang2019deep, li2019automatic, xu2020stylization, bai2022semantic, sheng2023guided}, text-based colorization \cite{bahng2018coloring, zou2019language, chang2022coder}) and fully-automatic colorization \cite{cheng2015deep, larsson2016learning, zhang2016colorful, iizuka2016let, isola2017image, royer2017probabilistic, guadarrama2017pixcolor, zhao2020pixelated, vitoria2020chromagan, su2020instance, zhao2020scgan, kumar2021colorization, wu2021towards, pucci2021collaborative, kong2021adversarial, saharia2022palette, kim2022bigcolor, huang2022unicolor}. Reference-based methods require additional user inputs that contain information relevant to the desired colors. The colorization systems use such information to assign possible colors to grayscale input images. Specifically, scribble-based methods propagate user-given color scribbles to the rest of the image. Exemplar-based methods attach colors from exemplar images to grayscale images. Text-based methods translate the information from words or languages into colors. Recently, deep neural networks improve colorization performance thanks to their superior feature representation ability. Built upon them, fully-automatic methods directly learn end-to-end colorization without any additional information on large datasets. To further improve the colorization quality, some specific designs have been used such as hyper-column \cite{larsson2016learning}, pre-trained backbones \cite{larsson2016learning, zhang2016colorful, iizuka2016let, vitoria2020chromagan, zhao2020scgan}, multi-task learning \cite{iizuka2016let, vitoria2020chromagan, zhao2020pixelated, zhao2020scgan, su2020instance}, and auxiliary loss functions \cite{isola2017image}.

\noindent \textbf{Video Colorization.} There are mainly four categories of existing video colorization methods in terms of data flow, as shown in Figure \ref{compare_arch_video_colorization}. The descriptions are as follows:

1) Image colorization and temporal smoothing \cite{bonneel2015blind, lai2018, lei2020blind, lei2022deep, lei2023blind} (Figure \ref{compare_arch_video_colorization} (a)): Based on the progress of image colorization methods, they added the temporal consistency on single colorized frames by post-processing them;

2) Image colorization (e.g., the first frame) and color propagation \cite{jampani2017video, meyer2018deep, endo2020practical, wu2020memory} (Figure \ref{compare_arch_video_colorization} (b)): They use one colorized frame to start the colorization. The colors and the correspondence of the following frames are learned sequentially;

3) Fully-automatic video colorization \cite{lei2019fully, kouzouglidis2019automatic, zhao2022vcgan} (Figure \ref{compare_arch_video_colorization} (c)): They learn the video colorization and temporal correspondence jointly on large video datasets by neural networks instead of learning it individually as categories 1) and 2);

4) Exemplar-based video colorization \cite{zhang2019deep, iizuka2019deepremaster, wan2022bringing} (Figure \ref{compare_arch_video_colorization} (d)): They propagate the colors from the exemplar image to the monochrome frames in a video.

Normally, the performance of both categories 1) and 2) are not satisfactory since the optimization of image colorization and temporal smoothing is separated. They require a powerful pre-trained image colorization algorithm, but normally the colorized videos are temporally not consistent. To address the issue, the other two categories combine image colorization and temporal smoothing together and learn them jointly. For instance, Lei \emph{et al.} \cite{lei2019fully} proposed a two-stage multi-modal video colorization framework. Kouzouglidis \emph{et al.} \cite{kouzouglidis2019automatic} adopted 3D convolution as the basic operator. Zhao \emph{et al.} \cite{zhao2022vcgan} used a global feature extractor and a placeholder feature extractor in the generator. Though the methods generate colorful videos automatically, their results are still not vivid enough. To improve the colorization quality, category 4) adopted an additional exemplar guidance image, e.g., Zhang \emph{et al.} \cite{zhang2019deep} aligned the exemplar with grayscale frames by a Nonlocal network and then fuse them by a ColorNet. It requires a high-quality exemplar image to obtain satisfactory results. To further ease the exemplar selection, we propose the first scribble-based video colorization method called SVCNet. The data flow of SVCNet is shown in Figure \ref{compare_arch_video_colorization} (e).

\noindent \textbf{Scribble-based Colorization.} The scribble-based colorization aims to propagate colors from user-given color scribbles to monochrome images. Common propagation schemes use local correspondences \cite{levin2004colorization}, edges \cite{huang2005adaptive}, or luminance-weighted chrominance blending \cite{yatziv2006fast}. However, these methods focused on local relations and failed to colorize the pixels far to color scribbles. To model the long-range connection, Xu \emph{et al.} \cite{xu2013sparse} proposed an affinity-based image editing scheme and Chen \emph{et al.} \cite{chen2012manifold} learned the mapping in feature space. However, the results are still highly related to the number or the location of given color scribbles. Moreover, the color bleeding effect is obvious when given color scribbles are close to the edges of objects. Recently, Zhang \emph{et al.} \cite{zhang2017real} used a neural network to extract the semantics and achieved better performance. In this paper, we further extend the scribble-based colorization to videos by the SVCNet framework.

\noindent \textbf{Saliency Detection.} It aims to localize the potential perceptual significant regions of the image by ``saliency map''. The early saliency detection methods were based on hand-crafted features such as color variation \cite{cheng2014global}, boundaries \cite{yang2013saliency}, and super-pixel \cite{li2015visual}, which predicted credible boundaries but not accurate structures of salient objects. Recent deep-learning-based methods generalized the saliency detection to diverse images and adopted different architectures such as recurrent network \cite{wang2016saliency}, encoder-decoder \cite{Liu2016DHSNet, Zhang2018Progressive, Liu2019PoolSal, wu2022edn}, and feature pyramid network \cite{wang2015deep, Zhao2019Pyramid, zhang2020multistage, siris2021scene} to fuse details and high-level semantics.

\noindent \textbf{Semantic Segmentation.} It aims to localize different objects in an image. The pioneer deep-learning-based method FCN \cite{long2015fully} used deconvolution and fusion of pooling layers. To enhance feature representation and context information, PSPNet \cite{zhao2017pyramid} adopted a pyramid pooling module and DeepLab \cite{chen2017deeplab} used different rates of dilated convolution. Bulti upon it, Chen \emph{et al}. \cite{chen2018encoder} combined the ASPPM with encoder-decoder architecture in DeepLab to let the network capture features in both cross and intra layers. Yang \emph{et al}. \cite{yang2018denseaspp} proposed a DenseASPPM to assemble different dilated branches. Recently, visual recognition in other data modalities such as 3D segmentation \cite{liu2022rm3d, liu2023fac, liu2022weakly} has achieved decent performance.

\begin{figure*}[t]
\begin{center}
\centering
\includegraphics[width=0.95\linewidth]{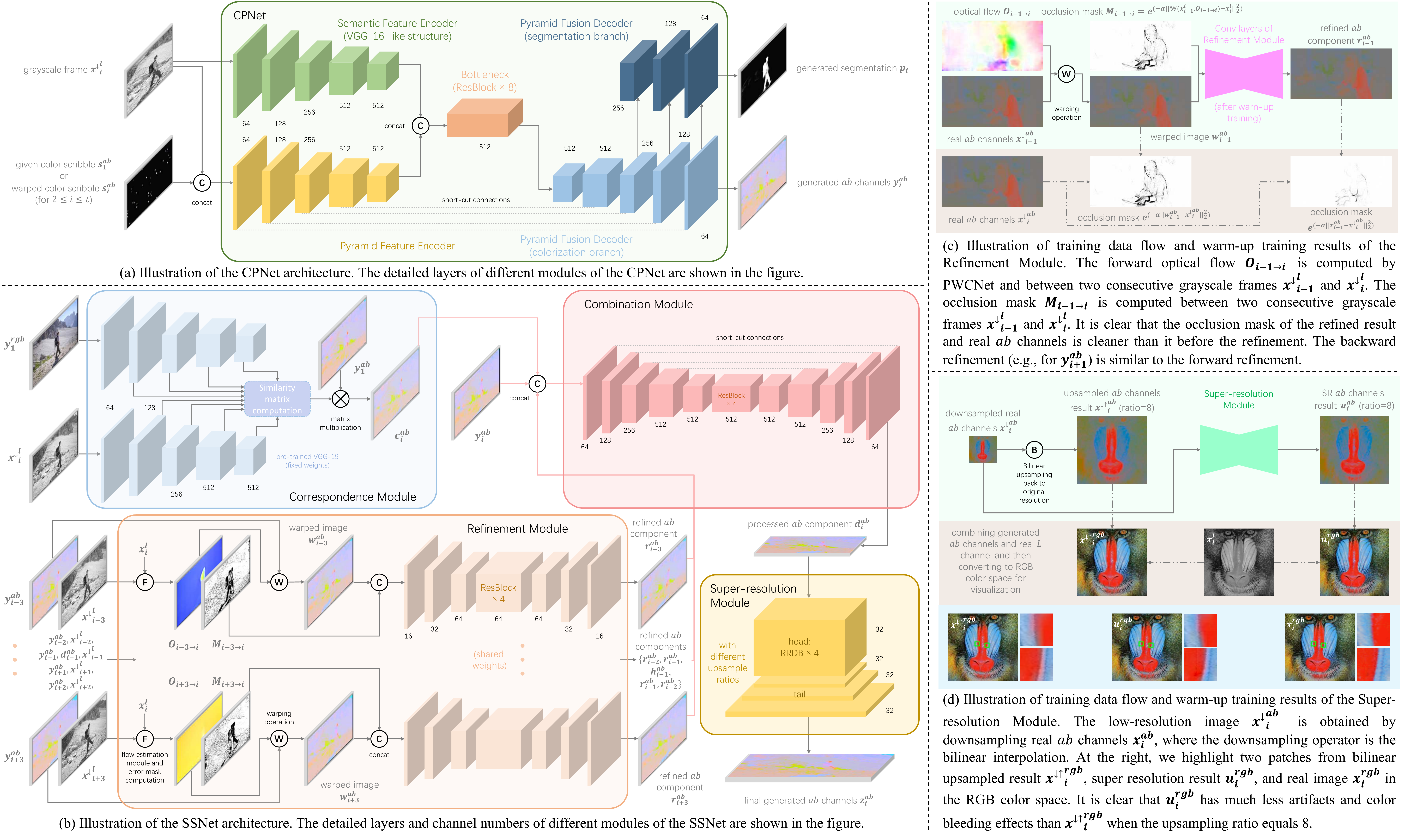}
\caption{Illustration of the detailed architectures for (a) CPNet and (b) SSNet. The special blocks are annotated while the other blocks are normal convolutional layers. The number of channels is annotated beside the blocks. Illustration of the training data flow and warm-up training results of (c) SSNet Refinement Module and (d) SSNet Super-resolution Module.}
\label{net}
\end{center}
\end{figure*}

\section{Methodology}

\subsection{Problem Formulation}

Given continuous grayscale frames, we aim to generate realistic color embeddings based on user-given color scribbles. To assist the colorization, we also generate corresponding segmentation maps to minimize color bleeding artifacts. We formulate the problem as maximizing a posteriori of color embeddings and segmentation maps conditioned on the grayscale frames $x_1^l, x_2^l, ..., x_t^l$, user-given color scribbles $s_1^{ab}$, and network parameters $\Theta$ of the SVCNet:
\begin{equation}
\Theta^* = \mathop{\arg\max}\limits_{\Theta} p( x_{1:t}^{ab}, q_{1:t} | x_{1:t}^l, s_{1}^{ab}, \Theta ),
\label{pf1}
\end{equation}
where $x_{1:t}^{l}$ and $x_{1:t}^{ab}$ are the grayscale component and color component in the \emph{CIE Lab} color space from the $1$-st frame to the $t$-th frame. $q_{1:t}$ are ground truth segmentation maps. $\Theta^*$ are theoretically optimal SVCNet parameters. To optimize the network, we further formulate the loss function $L$ on both color embeddings and segmentation maps:
\begin{equation}
\mathop{minimize}\limits_{SVCNet(.;\Theta)} \sum_{i=1}^{t} (L( z_{i}^{ab}, x_{i}^{ab}) + L( p_{i}, q_{i})),
\label{pf2}
\end{equation}
where $z_{i}^{ab}$ and $p_{i}$ are the outputs generated by the SVCNet at time step $i$. $x_{i}^{ab}$ and $q_{i}$ are ground truth color embeddings and segmentation maps at time step $i$. $SVCNet(.;\Theta)$ denotes the SVCNet with network parameter $\Theta$.

\subsection{SVCNet Architecture}

The SVCNet consists of two sub-networks: color propagation sub-network (CPNet) and spatiotemporal smoothing sub-network (SSNet), as shown in Figures \ref{pipeline} and \ref{net}. Below we present every sub-network and module.

\noindent \textbf{CPNet}. CPNet performs scribble-based image colorization, which includes two encoders and two decoders, as shown in Figure \ref{net} (a). To merge the information of grayscale input and color scribble guidance, we adopt a pyramid feature encoder. Considering that the color scribbles for far time steps are obtained by warping the first color scribble may vanish, we strengthen the colorization quality by the other semantic feature encoder, which is a pre-trained network. The resulting features are concatenated and processed by a bottleneck for fusion. Afterward, the pyramid feature decoder produces a color embedding from the output of bottleneck and short-cut connections of the pyramid feature encoder \cite{ronneberger2015u}. In addition, the last three decoder layers are fed into a segmentation branch to produce corresponding segmentation maps. At the training, the segmentation branch helps revise the weights of the colorization branch through backpropagation.

\noindent \textbf{SSNet}. SSNet post-processes the output of CPNet. It enhances the temporal consistency of generated frames and includes four modules, as shown in Figure \ref{net} (b).

\noindent \textbf{Refinement Module}. It is designed for processing short-range correlations, which are modeled by both previous frames ($y_{i-3}^{ab}, y_{i-2}^{ab}, y_{i-1}^{ab}, d_{i-1}^{ab}$) and leading frames ($y_{i+1}^{ab}, y_{i+2}^{ab}, y_{i+3}^{ab}$) on time step $i$. To match their locations with the current frame, we first warp them to the position of the current frame with optical flows, which are computed by a pre-trained PWC-Net \cite{sun2018pwc} on grayscale frames. Then, we use the Refinement Module to post-process them in order to minimize the warping artifacts caused by occlusions and motion boundaries. To represent such regions, we compute the occlusion mask for warped previous and leading frames, which are defined as:
\begin{equation}
M_{j \rightarrow i} = exp (- \alpha || \mathbb{W} ( {x^{\downarrow}}_{j}^l, O_{j \rightarrow i} ) - {x^{\downarrow}}_i^l ||^2_2 ),
\label{ssnet1}
\end{equation}
where $M_{j \rightarrow i}$ and $O_{j \rightarrow i}$ denote the occlusion mask and the optical flow from time step $j$ to $i$, respectively. ${x^{\downarrow}}_{j}^l$ and ${x^{\downarrow}}_{i}^l$ are downsampled real grayscale images at time steps $j$ and $i$, respectively. $\mathbb{W}(*)$ is the warping operation. $\alpha$ is a hyper-parameter that controls the sensitivity of the occlusion mask. The forwarding procedure of Refinement Module ($\rm RM$) can be represented as:
\begin{equation}
r_j^{ab} = {\rm RM} (\mathbb{W} ( y_{j}^{ab}, O_{j \rightarrow i} ), M_{j \rightarrow i} ),
\label{ssnet2}
\end{equation}
where we set $j = (i-3, i-2, i-1, i+1, i+2, i+3)$. Also, we refine the previous SSNet low-resolution output $d_{i-1}^{ab}$, which is the output of the Combination Module, as follows:
\begin{equation}
h_{i-1}^{ab} = {\rm RM} (\mathbb{W} ( d_{i-1}^{ab}, O_{i-1 \rightarrow i} ), M_{i-1 \rightarrow i} ).
\label{ssnet3}
\end{equation}

\noindent \textbf{Correspondence Module}. The long-range connection between the first colorized frame and the current frame is modeled by the Correspondence Module. We do not use the optical flow since there often exist large motions between them. On the contrary, we compute the similarity matrix between them and use it to warp the features of the first generated color embeddings. Firstly, like in \cite{zhang2019deep}, we build the feature pyramid of RGB-formatted first frame $y_1^{rgb}$ and current grayscale frame ${x^{\downarrow}}_{i}^l$ ($y_1^{rgb}, {x^{\downarrow}}_{i}^l \in \mathbb{R}^{H \times W \times C}$) by extracting their features of layers $conv_{2\_2}$, $conv_{3\_2}$, $conv_{4\_2}$, and $conv_{5\_2}$ of a pre-trained VGG-19 network \cite{simonyan2014very}. All the features are upsampled to the same resolution as $conv_{2\_2}$ (i.e., $\frac{1}{2}H \times \frac{1}{2}W$) and then concatenated together. The resulting features for $y_1^{rgb}$ and ${x^{\downarrow}}_{i}^l$ are denoted as $F_1$ and $F_i$, respectively. Secondly, the similarity matrix $S_{1 \leftrightarrow i} \in \mathbb{R}^{\frac{1}{4}HW \times \frac{1}{4}HW}$ is computed as:
\begin{equation}
S_{1 \leftrightarrow i} = \frac{ F_1 - \mu(F_1) }{ || F_1 - \mu(F_1) ||_2 } \cdot \frac{ F_i - \mu(F_i) }{ || F_i - \mu(F_i) ||_2 } ,
\label{ssnet4}
\end{equation}
where $\mu(*)$ denotes the mean operation. Then, we use the similarity matrix to warp $y_1^{ab}$ \cite{zhang2019deep} as follows:
\begin{equation}
c_i^{ab} = ( \sum\limits_m \mathop{softmax} \limits_h ( \tau \cdot S_{1 \leftrightarrow t} (:, m) ) \cdot {y^{\downarrow}}_1^{ab} )^{\uparrow} ,
\label{ssnet5}
\end{equation}
where $c_i^{ab}$ is the warped result of the first generated color embeddings $y_1^{ab}$. $\tau$ is the temperature parameter. There is a downsampling operation since we need to match the resolution of the similarity map and the color embeddings.

\noindent \textbf{Combination Module}. Next, the Combination Module aggregates the information from outputs of the Refinement Module ($r_{i-3}^{ab}, r_{i-2}^{ab}, r_{i-1}^{ab}, h_{i-1}^{ab}, r_{i+1}^{ab}, r_{i+2}^{ab}, r_{i+3}^{ab}$), Correspondence Module ($c_i^{ab}$), and the current generated color embeddings from the CPNet ($y_i^{ab}$) by a U-Net-like architecture \cite{ronneberger2015u}. The output of the Combination Module is denoted as $d_i^{ab}$.

\noindent \textbf{Super-resolution Module}. Finally, since previous operations run on a fixed small resolution, the Super-resolution Module recovers the frames with original sizes at inference from $d_i^{ab}$. For different high resolutions, we use the same feature extraction head but different tails with multiple upsampling ratios. The final output is denoted as $z_i^{ab}$.

\subsection{Training Strategy}

Directly optimizing the large SVCNet without any initialization easily encounters the gradient exploding issue. To stabilize the training, we propose the warm-up pre-training for CPNet and some modules of SSNet, respectively. After that, we train the full SVCNet on video datasets. The details are as follows:

1) CPNet warm-up pre-training: We pre-train the CPNet on ImageNet dataset \cite{russakovsky2015imagenet}, which provides much more modes and scenes than video datasets. Since the ImageNet dataset does not provide segmentation maps, we generate saliency maps as ground truth by \cite{Zhao2019Pyramid}. We then finetune the CPNet on single frames from video datasets (DAVIS \cite{perazzi2016benchmark} and Videvo \cite{Videvo}). It makes CPNet fit the sizes of video frames better. Similarly, we generate saliency maps as pseudo segmentation maps for the Videvo dataset by \cite{Zhao2019Pyramid}.

2) SSNet warm-up pre-training: We conduct the self-supervised learning for the Refinement Module and the Super-resolution Module on two video datasets: DAVIS \cite{perazzi2016benchmark} and Videvo \cite{Videvo}. Firstly, for a frame $x_i^{ab}$ in a video, we stochastically warp the previous frame $x_{i-1}^{ab}$ or leading frame $x_{i+1}^{ab}$ to the position of $x_t^{ab}$ as the input for the Refinement Module. We make it reconstruct itself and train it with an $L1$ loss. Secondly, we downsample a frame of the original resolution with different ratios (2, 4, and 8) as the input for the Super-resolution Module. We then let the Super-resolution Module reconstruct itself and train it with an $L1$ loss. The warm-up training is conducted only on \emph{ab} color components. It ends when the loss is small and stable enough. The results are shown in Figure \ref{net} (c) and (d). The Refinement Module can reduce the artifacts in the occluded regions. The Super-resolution Module can minimize the color bleeding artifacts especially when the upsampling ratio is large (e.g., 8).

3) joint training stage: We optimize the full SVCNet (CPNet and SSNet) based on all the warm-up training weights. We randomly select 7 continuous frames in a video and then flip the first 6 frames as a batch at the training. In what follows, we will introduce the loss functions for the ``CPNet warm-up pre-training'' and ``joint training stage''.

\begin{table*}[t]
\begin{center}
\caption{Conclusion of optimization details for different training stages.}
\label{training_strategy}
\begin{tabular}{ccccccc}
\hline
Training stage & Trained network & Loss & Training set & Total iterations & Initial learning rate (LR) & LR halved iteration \\
\hline
1) CPNet warm-up & CPNet & $L^{CP}$ & ImageNet & 800,740 & 1$\times$10$^{-4}$ & 400,370 \cr pre-training & CPNet & $L^{CP}$ & DAVIS+Videvo & 200,600 & 5$\times$10$^{-5}$ & 100,300 \\
\hline
2) SSNet warm-up & Refinement Module & $L1$ & DAVIS+Videvo & 312,000 & 1$\times$10$^{-4}$ & 156,000 \cr pre-training & \scriptsize{Super-resolution Module} & $L1$ & DAVIS+Videvo & 296,200 & 1$\times$10$^{-4}$ & 148,100 \\
\hline
3) joint training stage & CPNet+SSNet & $L^{joint}$ & DAVIS+Videvo & 312,000 & \scriptsize{CPNet:1$\times$10$^{-6}$; SSNet:5$\times$10$^{-5}$} & 156,000 \\
\hline
\end{tabular}
\end{center}
\end{table*}

\subsection{Loss Function}

When training CPNet, we apply a colorization loss and a segmentation loss for the outputs of the two pyramid fusion decoder branches, respectively:
\begin{equation}
L_c^{CP} = \mathbb{E} [|| y_i^{ab} - {x^{\downarrow}}_i^{ab} ||_1 ],
\label{loss_cpnet_1}
\end{equation}
\begin{equation}
L_s^{CP} = \mathbb{E} [|| p_i - q_i ||_1 ],
\label{loss_cpnet_2}
\end{equation}
where $y_i^{ab}$ and $p_i$ are the outputs of two branches. ${x^{\downarrow}}_i^{ab}$ and $q_i$ are the corresponding ground truth. Then, the total loss function for the CPNet warm-up training is defined as:
\begin{equation}
L^{CP} = L_c^{CP} + \lambda_s L_s^{CP},
\label{loss_cpnet}
\end{equation}
where $\lambda_s$ is the trade-off parameter.

For the joint training, we maintain the CPNet loss function and add the SSNet loss function. Specifically, we adopt colorization losses for the output of the Combination Module ($d_i^{ab}$) and the output of the Super-resolution Module ($z_i^{ab}$), respectively. They can be represented as:
\begin{equation}
L^{SS} = \mathbb{E} [|| d_i^{ab} - {x^{\downarrow}}_i^{ab} ||_1 ] + \mathbb{E} [|| z_i^{ab} - x_i^{ab} ||_1 ],
\label{loss_ssnet_1}
\end{equation}
where ${x^{\downarrow}}_i^{ab}$ and $x_i^{ab}$ are ground truth at low resolution and original resolution, respectively.

The full loss function of the joint training stage is the sum:
\begin{equation}
L^{joint} = L^{CP} + L^{SS}.
\label{loss_ssnet}
\end{equation}

\section{Experiment}

\subsection{Implementation Details}

\noindent \textbf{Training Dataset.} We use the entire ImageNet dataset for the CPNet warm-up pre-training. It includes 1.3 million images with 1000 categories. We use DAVIS and Videvo datasets for other training stages. They include 156 short video clips with 29620 frames. The DAVIS dataset has labeled binary segmentation maps. For ImageNet and Videvo datasets, we generate saliency maps as pseudo-segmentation maps. As for pre-processing, ImageNet images are resized to the resolution of 256$\times$256. The video frames from DAVIS and Videvo datasets are resized to 256$\times$448 as the input of the SVCNet and 512$\times$896 as the ground truth (i.e., the upsampling ratio for the Super-resolution Module is set to 2 at the joint training stage). All images are normalized to the range of [0, 1].

\noindent \textbf{Color Scribble.} We randomly select color scribbles from \emph{ab} channels of ground truth images. There is a half probability to use valid color scribbles in the training and the number of color scribbles ranges from 1 to 40. At the joint training stage, color scribbles are only provided for the first frame in a batch. The following color scribbles are obtained by warping the previous one with forward optical flows.

\noindent \textbf{Optimization.} The optimization details are concluded in Table \ref{training_strategy}, where we list trained networks, loss functions, training sets, total training iterations, initial learning rates (LRs), and the specific iterations when learning rates halved, respectively, in every training stage. At the start of the joint training stage, we initialize both CPNet and SSNet with warm-up pre-training weights. For the remaining layers or blocks, we initialize them by \cite{glorot2010understanding}. The batch size for the warm-up pre-training stages is 4 and it is 1 for the joint training stage per GPU. We use the Adam optimizer \cite{kingma2014adam} with $\beta_1$=0.5, $\beta_2$=0.999. The trade-off parameter $\lambda_s$ is set to 0.1. Both the mask parameter $\alpha$ and temperature parameter $\tau$ are 200. We implement the SVCNet with the PyTorch framework. We train it on 8 NVIDIA V100 GPUs and 8 NVIDIA Titan Xp GPUs. Considering the parallel training, it takes approximately 10 days to complete the warm-up pre-training and another 6 days to complete the joint training.

\noindent \textbf{Network Architecture.} The SVCNet architecture is shown in Figure \ref{net}, where we emphasize special blocks such as short-cut connections \cite{ronneberger2015u}, residual block (ResBlock) \cite{he2016deep}, and residual-in-residual dense block (RRDB) \cite{wang2018esrgan}. We also emphasize the number of channels. We use LeakyReLU \cite{maas2013rectifier} as the activation function except for the first and second layers. We use instance normalization \cite{ulyanov2016instance} only in the Refinement Module.

\subsection{Quantitative Metrics}

\noindent \textbf{Generation Quality.} We adopt PSNR and SSIM \cite{wang2004image} to calculate pixel-level accuracy and structural similarity between generated results and ground truth, respectively. For the scribble-based colorization task, some ground truth scribbles are given in the validation stage and the colorization becomes a specific task. Therefore, PSNR and SSIM are proper metrics to evaluate the generation quality.

\noindent \textbf{Segmentation Performance.} We use HRNetV2 + OCR \cite{sun2019deep, yuan2019segmentation} to calculate the mean intersection over union (mIoU) of the generated frames on DAVIS semantic segmentation validation dataset \cite{perazzi2016benchmark}. If a method obtains a higher mIoU value, it may have less probability to encounter the color bleeding issue since the segmentation algorithm can better separate key objects from the colorized image of this method.

\noindent \textbf{Human Preference.} We conduct a human preference study on video colorization results from different methods. There are 10 videos randomly selected from DAVIS and Videvo validation sets for users to compare. For each video, the human observer needs to select the best result based on temporal consistency and color vividness. There are 10 human observers in the experiment and they can watch the videos many times.

\begin{figure*}[t]
\begin{center}
\centering
\includegraphics[width=0.95\linewidth]{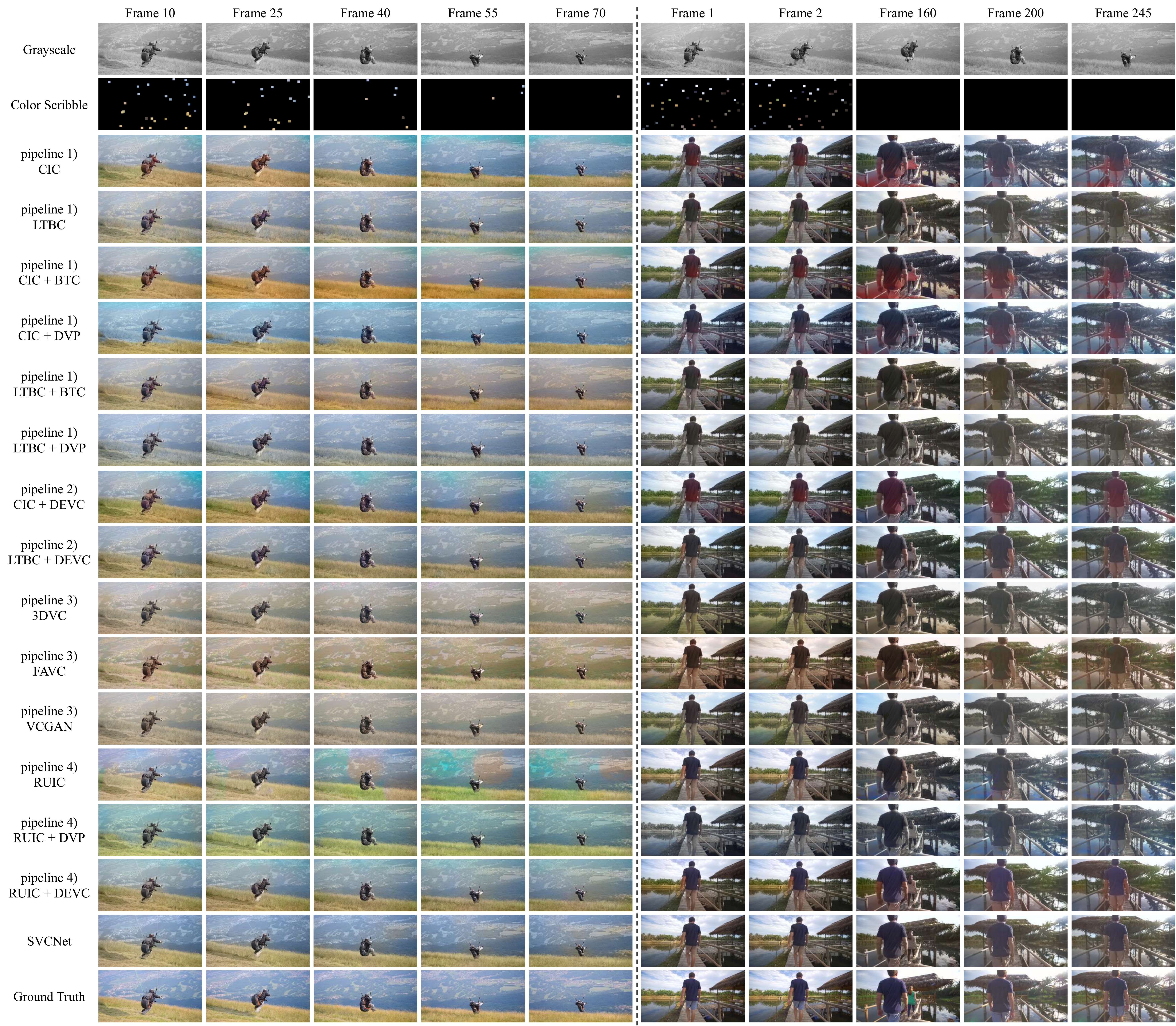}
\caption{Illustration of video colorization results of SVCNet and other pipelines. The images are extracted from the ``paragliding-launch'' sample of the DAVIS dataset and the ``YogaHut2'' sample of the Videvo dataset, respectively.}
\label{video}
\end{center}
\end{figure*}

\begin{figure*}[t]
\begin{center}
\centering
\includegraphics[width=0.95\linewidth]{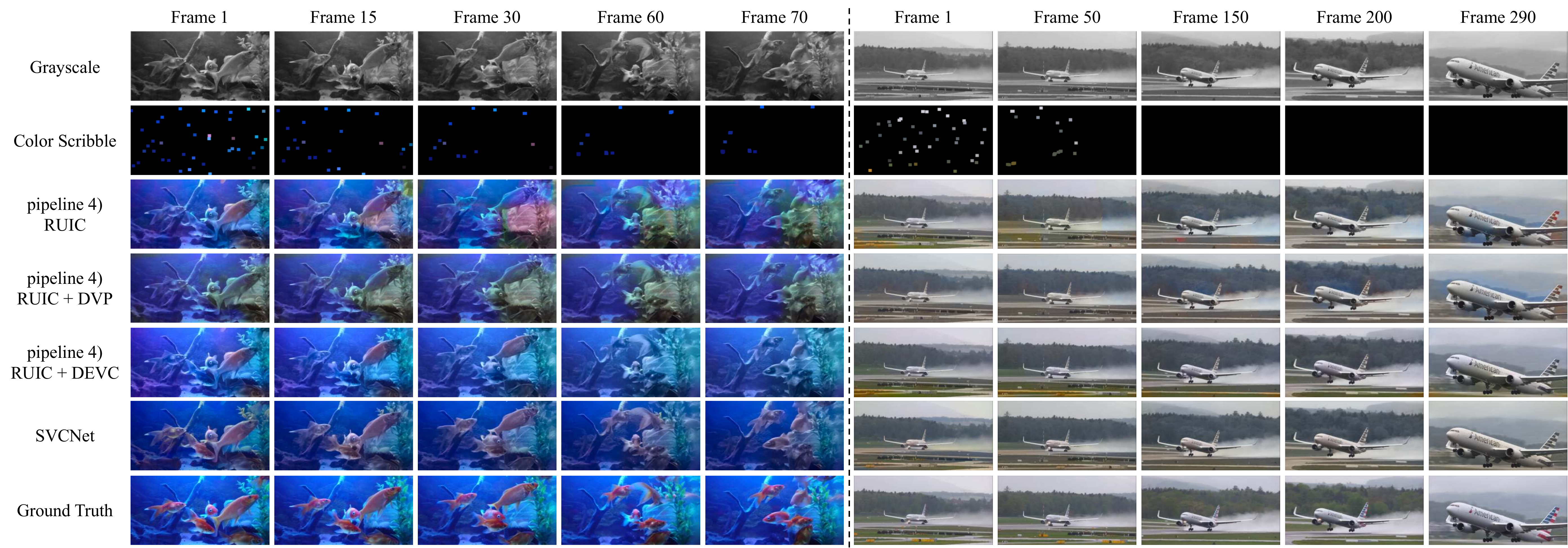}
\caption{Illustration of video colorization results of SVCNet and state-of-the-art scribble-based video colorization methods. The images are extracted from the ``gold-fish'' sample of the DAVIS dataset and the ``AircraftTakingOff1'' sample of the Videvo dataset, respectively.}
\label{video2}
\end{center}
\end{figure*}

\begin{figtab*}
\hspace{1mm}
\begin{minipage}[b]{0.7\linewidth}
\centering
\tabcaption{Comparison of video colorization pipelines and the proposed SVCNet on DAVIS and Videvo datasets. The \textcolor{red}{\textbf{red}}, \textcolor{blue}{\textbf{blue}}, and \textcolor{green}{green} colors represent the best, the second-best, and the third-best performance, respectively. The ``IC'' and ``VC'' represent the ``image colorization method'' and ``video colorization method'', respectively.}
\label{table_sota}
\resizebox{124mm}{43mm}{
\begin{tabular}{lccccccc}
\hline
\multirow{2}{*}{Method} & \multirow{2}{*}{Type} & Color & \multicolumn{3}{c}{DAVIS} & \multicolumn{2}{c}{Videvo} \cr & & scribble & PSNR $\uparrow$ & SSIM $\uparrow$ & mIoU $\uparrow$ & PSNR $\uparrow$ & SSIM $\uparrow$ \\
\hline
1) CIC \cite{zhang2016colorful} & IC & / & 22.48 & 0.9043 & 0.8419 & 21.83 & 0.9024 \\
1) LTBC \cite{iizuka2016let} & IC & / & 23.96 & 0.9171 & 0.8945 & 24.69 & 0.9272 \\
1) ChromaGAN \cite{vitoria2020chromagan} & IC & / & 23.77 & 0.9420 & 0.8879 & 23.94 & 0.9392 \\
1) IAC \cite{su2020instance} & IC & / & 22.88 & 0.9422 & 0.8836 & 23.99 & 0.9486 \\
1) CIC + BTC \cite{zhang2016colorful, lai2018} & VC & / & 21.50 & 0.8935 & 0.8419 & 21.05 & 0.8833 \\
1) LTBC + BTC \cite{iizuka2016let, lai2018} & VC & / & 22.47 & 0.9044 & 0.8899 & 22.83 & 0.9105 \\
1) ChromaGAN + BTC \cite{vitoria2020chromagan, lai2018} & VC & / & 19.91 & 0.8938 & 0.8796 & 16.64 & 0.8325 \\
1) IAC + BTC \cite{su2020instance, lai2018} & VC & / & 18.40 & 0.8778 & 0.8698 & 15.85 & 0.8209 \\
1) CIC + DVP \cite{zhang2016colorful, lei2020blind} & VC & / & 23.30 & 0.9351 & 0.8724 & 22.23 & 0.9328 \\
1) LTBC + DVP \cite{iizuka2016let, lei2020blind} & VC & / & 24.06 & 0.9425 & 0.8886 & 24.75 & \textcolor{green}{0.9548} \\
1) ChromaGAN + DVP \cite{vitoria2020chromagan, lei2020blind} & VC & / & 23.81 & 0.9444 & 0.8800 & 23.97 & 0.9451 \\
1) IAC + DVP \cite{su2020instance, lei2020blind} & VC & / & 22.91 & 0.9407 & 0.8936 & 23.99 & 0.9503 \\
2) CIC + DEVC \cite{zhang2016colorful, zhang2019deep} & VC & / & 21.64 & 0.9321 & 0.8661 & 21.36 & 0.9231 \\
2) LTBC + DEVC \cite{iizuka2016let, zhang2019deep} & VC & / & 22.46 & 0.9397 & 0.8688 & 24.03 & 0.9513 \\
2) ChromaGAN + DEVC \cite{vitoria2020chromagan, zhang2019deep} & VC & / & 22.56 & 0.9429 & \textcolor{green}{0.9026} & 22.40 & 0.9386 \\
2) IAC + DEVC \cite{su2020instance, zhang2019deep} & VC & / & 22.43 & 0.9384 & 0.8712 & 23.59 & 0.9479 \\
3) 3DVC \cite{kouzouglidis2019automatic} & VC & / & 23.49 & 0.9151 & 0.8948 & 24.33 & 0.9231 \\
3) FAVC \cite{lei2019fully} & VC & / & 22.98 & 0.9055 & 0.8889 & 23.47 & 0.9183 \\
3) VCGAN \cite{zhao2022vcgan} & VC & / & 23.43 & 0.9133 & 0.8954 & 24.73 & 0.9225 \\
\hline
4) RUIC \cite{zhang2017real} & IC & \checkmark & \textcolor{green}{25.42} & \textcolor{green}{0.9456} & 0.8995 & \textcolor{green}{25.02} & 0.9432 \\
4) RUIC + BTC \cite{zhang2017real, lai2018} & VC & \checkmark & 21.16 & 0.8869 & 0.8900 & 17.07 & 0.8279 \\
4) RUIC + DVP \cite{zhang2017real, lei2020blind} & VC & \checkmark & \textcolor{red}{25.82} & 0.9455 & \textcolor{blue}{0.9075} & 24.68 & 0.9460 \\
4) RUIC + DEVC \cite{zhang2017real, zhang2019deep} & VC & \checkmark & 24.85 & \textcolor{blue}{0.9524} & 0.9002 & \textcolor{blue}{25.66} & \textcolor{blue}{0.9583} \\
\hline
SVCNet & VC & \checkmark & \textcolor{blue}{25.71} & \textcolor{red}{0.9565} & \textcolor{red}{0.9104} & \textcolor{red}{26.30} & \textcolor{red}{0.9615} \\
\hline
\end{tabular}
}
\end{minipage}
\quad
\begin{minipage}[h]{0.25\linewidth}
\centering
\includegraphics[width=\linewidth]{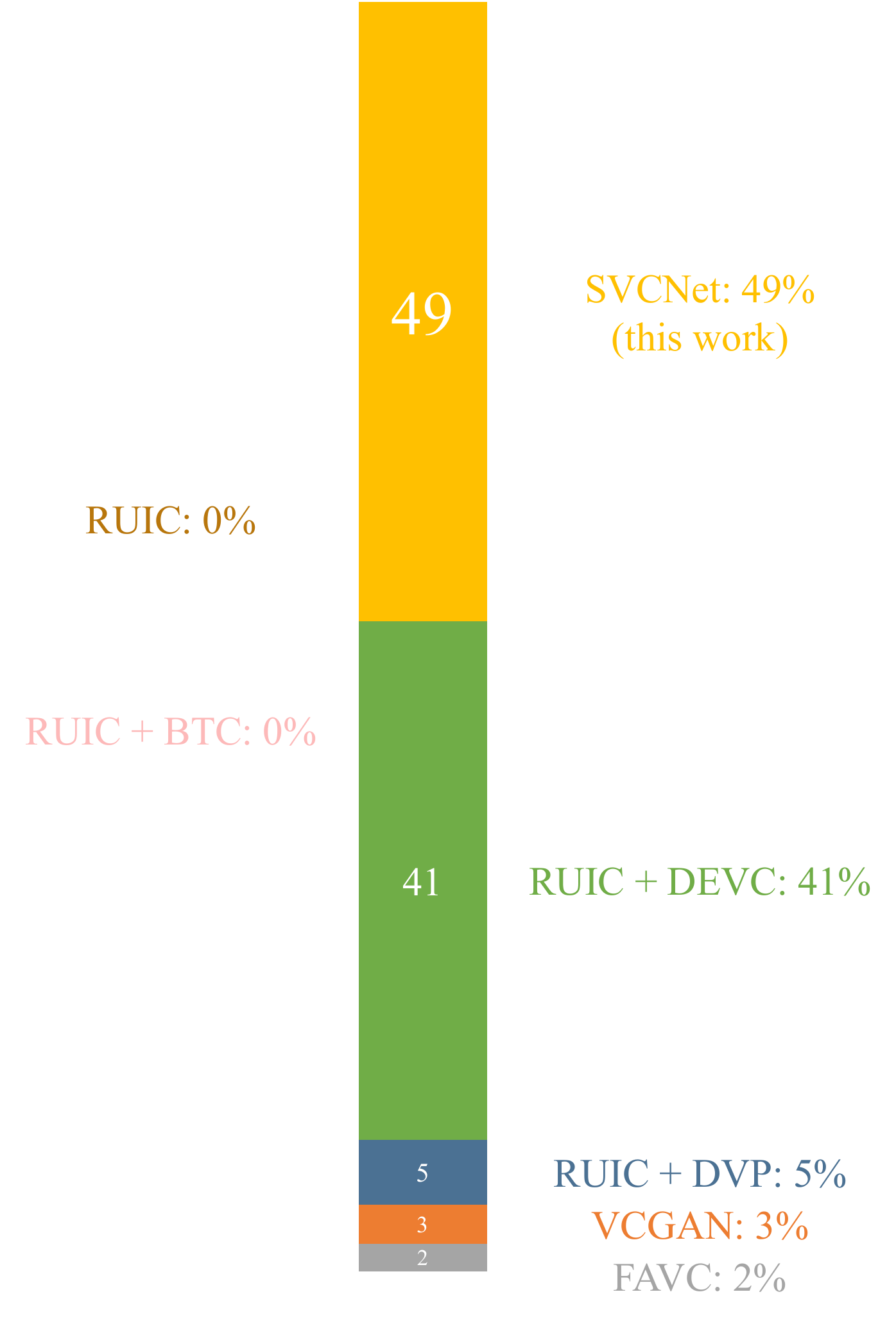}
\figcaption{Human preference study result. The human preference rate for each method is marked in the figure. Different colors denote different methods.}
\label{human_study}
\end{minipage}
\end{figtab*}

\begin{figure}[t]
\begin{center}
\centering
\includegraphics[width=0.95\linewidth]{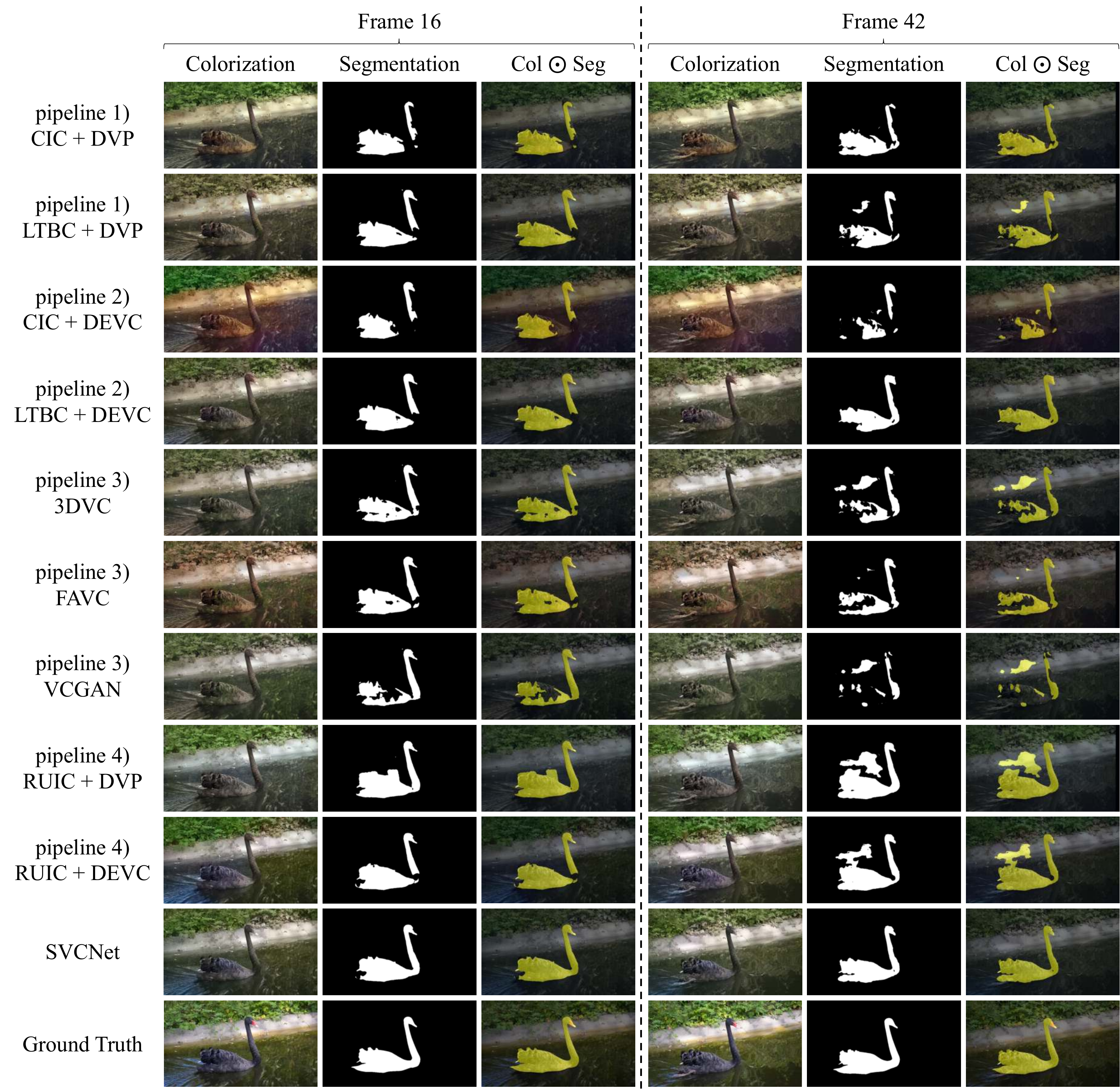}
\caption{Illustration of video colorization and segmentation results of SVCNet and other pipelines on the ``blackswan'' sample of the DAVIS dataset.}
\label{seg}
\end{center}
\end{figure}

\subsection{Video Colorization Experiments}

\noindent \textbf{Experiment Setting.} We compare the video colorization performance of SVCNet and other recent works with similar targets, which can be categorized into four pipelines:

1) Image colorization and temporal smoothing (as shown in Figure \ref{compare_arch_video_colorization} (a)): CIC \cite{zhang2016colorful}, LTBC \cite{iizuka2016let}, ChromaGAN \cite{vitoria2020chromagan}, and IAC \cite{su2020instance} are used as image colorization algorithms; BTC \cite{lai2018} and DVP \cite{lei2020blind} are used as temporal smoothing networks;

2) Image colorization and color propagation (as shown in Figure \ref{compare_arch_video_colorization} (b)): DEVC \cite{zhang2019deep} is used to propagate the first colorized frame to the following frames;

3) Fully-automatic video colorization (as shown in Figure \ref{compare_arch_video_colorization} (c)): FAVC \cite{lei2019fully}, 3DVC \cite{kouzouglidis2019automatic}, and VCGAN \cite{zhao2022vcgan};

4) Scribble-based image colorization and temporal smoothing/color propagation: RUIC \cite{zhang2017real} is used as the image scribble-based image colorization method; DVP \cite{lei2020blind} and DEVC \cite{zhang2019deep} are used as post-processing methods. Note that, RUIC is finetuned on video datasets.

We perform experiments on DAVIS \cite{perazzi2016benchmark} and Videvo \cite{Videvo} validation sets. There are overall 50 video clips and each of them contains approximately 100 frames. In the experiment, we use 40 color scribbles for scribble-based methods, which are randomly extracted from the first frame. The following scribbles are obtained by warping color scribbles of the first frame with forward optical flows. Different pipelines share the same color scribbles. For the SVCNet, we set the up-sampling ratio of the Super-resolution module as 2 to obtain results with a resolution of 512$\times$896 and then resize them to target resolutions of DAVIS and Videvo datasets. The other methods run on the same resolution as validation images. The overall tuning epoch of DVP is set to 25 as default. Note that, FAVC and DVP automatically crop input images to make side lengths a multiple of 32. Therefore, we only use valid regions to calculate quantitative metrics for them. The quantitative results are concluded in Table \ref{table_sota} and some samples are illustrated in Figures \ref{video} and \ref{video2}.

\noindent \textbf{Colorization Quality.} Firstly, from Figure \ref{video}, SVCNet results are more approach to the ground truth than the results of other methods in terms of the color tone and color naturalness. For instance, the results from 3DVC, FAVC, and VCGAN on the ``paragliding-launch'' sample are less colorful. These methods do not well balance the video colorization vividness and temporal smoothness, i.e., their results are too smooth so they are not colorful enough. Although BTC, DVP, and DEVC can smooth single-colorized frames, the results are still not reasonable. For instance, there are obvious color artifacts in CIC-related results on the ``YogaHut2'' sample, e.g., the clothes are colorized in red. Since CIC predicts a color distribution for each pixel in an image, the output is not always natural compared with ground truth.

Secondly, there is less color bleeding issue in SVCNet results than in other methods. For instance, the human is colorized to blue for CIC + DVP on the ``paragliding-launch'' sample since it cannot distinguish the human and the sky. The hand is colorized to green for 3DVC on the ``YogaHut2'' sample since the neighboring green colors of trees affect the colorization of it. Compared with other methods, SVCNet has much fewer color bleeding artifacts since it uses a segmentation branch to alleviate the problem.

Thirdly, SVCNet obtains better video temporal consistency than other methods since it aggregates both short-range connections and the long-range connection. It obtains better temporal consistency than other methods. For pipeline 1), the videos generated by image colorization and temporal smoothing methods are not continuous enough, e.g., CIC + BTC, CIC + DVP, LTBC + BTC, and LTBC + DVP. Since image colorization and temporal smoothing are learned individually, the final outputs are still close to the single-frame colorized results. For instance, the human is colorized to red in frame 160 for the CIC results while it turns to gray for frame 200; but the CIC + BTC cannot alleviate this issue, i.e., the results are not temporal consistent enough. For pipeline 2), the videos are smoother than the results from pipeline 1). However, it exists a similar problem since image colorization methods and DEVC are not jointly trained. For pipeline 3), the video continuity is good compared with other methods, but the results of pipeline 3) are less colorful. For pipeline 4), RUIC + DEVC outperforms the other two methods RUIC and RUIC + DVP. However, it remains the common problem of DEVC-based methods, i.e., results highly rely on the first colorized frame.

Finally, SVCNet better utilizes the given color scribbles than RUIC. Since only color scribbles of the first frame are given, they might vanish for far frames. In such circumstances, RUIC cannot produce appropriate results, e.g., there are obvious flickering artifacts in RUIC results on the ``paragliding-launch'' sample. Although DVP and DEVC can bring temporal consistency to the results from RUIC, they are not close enough to the ground truth. We illustrate more examples in Figure \ref{video2}, where results of RUIC + DVP and RUIC + DEVC are inferior to SVCNet in terms of color vividness and accuracy compared with ground truth.

\noindent \textbf{Colorization Fidelity.} According to Table \ref{table_sota}, SVCNet achieves better performance in terms of both PSNR and SSIM values than other methods. It demonstrates that SVCNet can well use the colors from given color scribbles. Compared with the state-of-the-art method RUIC + DEVC, SVCNet additionally adopts short-range connections in the SSNet; while compared with RUIC + DVP, SVCNet additionally uses the long-range connection. The short-range connections include the previous three and leading three single-frame colorization results and the last output. The long-range connection is the information from the first frame colorization result. Since SVCNet aggregates short-range information, long-range information, and the output of the current time step, it obtains higher colorization fidelity.

\noindent \textbf{Human Preference.} We select the top-performed methods FAVC, VCGAN, RUIC, RUIC + BTC, RUIC + DVP, RUIC + DEVC and the SVCNet in the human preference study. The results are shown in Figure \ref{human_study}. It is clear that the proposed SVCNet achieves a better human preference rate than other methods. The experiment demonstrates that colorization results from SVCNet are more temporally consistent and colorful compared with other results.

\begin{figure}[t]
\begin{center}
\centering
\includegraphics[width=0.95\linewidth]{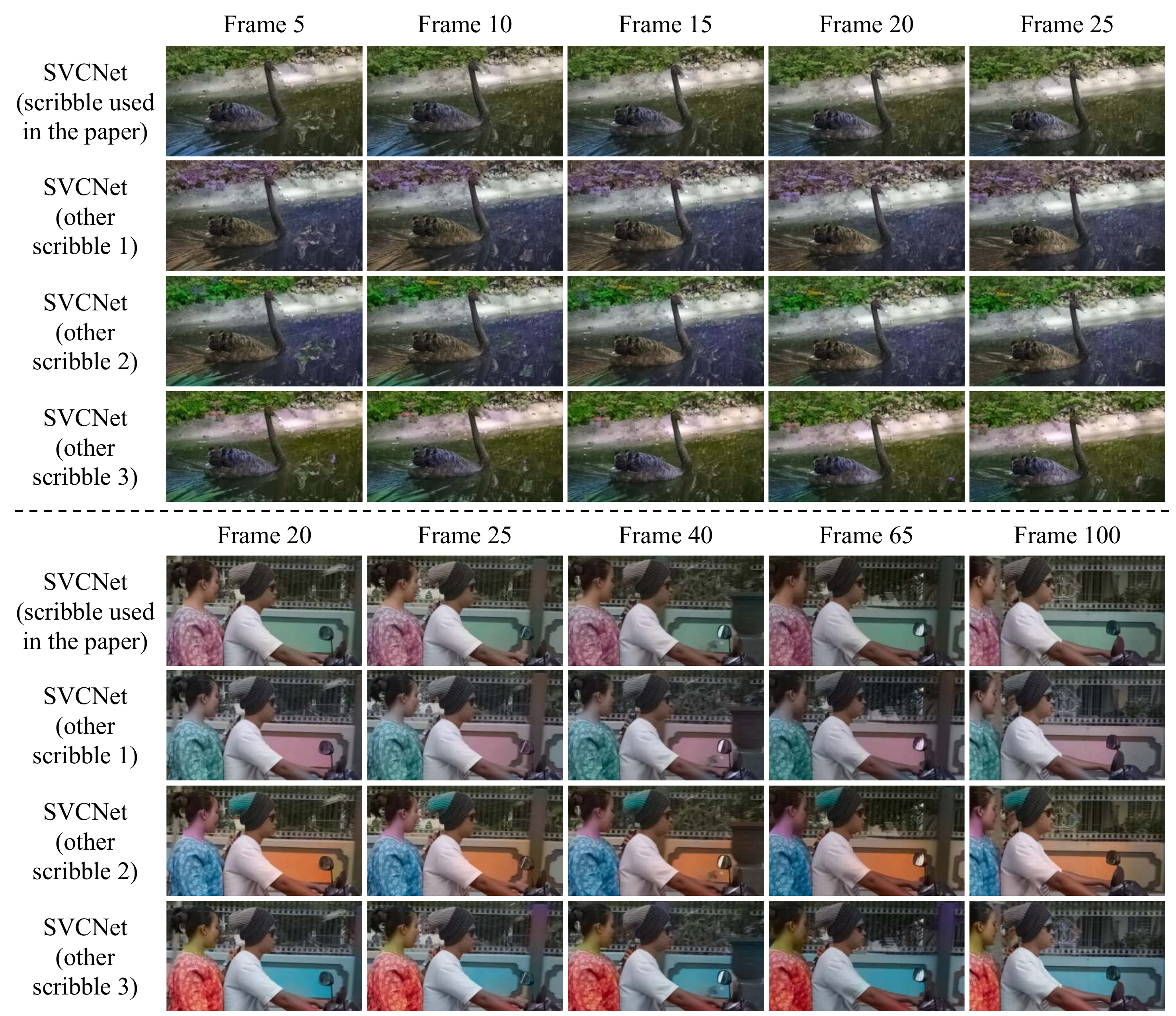}
\caption{Illustration of SVCNet results from different color scribbles. The images are extracted from the ``blackswan'' sample of the DAVIS dataset and the ``CoupleRidingMotorbike'' sample of the Videvo dataset, respectively. The input color scribbles are omitted.}
\label{diversity}
\end{center}
\end{figure}

\noindent \textbf{Color Bleeding Analysis.} We further discuss the color bleeding artifacts in this subsection. As shown in Figure \ref{seg}, the segmentation algorithm produces the clearest segmentation map from the SVCNet results. For instance, the contour of the swan is more continuous for the SVCNet than the other methods. It represents the key objects are more distinguishable from other objects for it. In addition, SVCNet achieves the highest mIoU among all the methods, as concluded in Table \ref{table_sota}. It demonstrates that the objects of its results are more obvious than the other methods, which denotes that SVCNet has less possibility to encounter color bleeding artifacts. This is achieved by adding an additional segmentation loss at the training, where the weights of the colorization branch are tuned by the segmentation branch.

\noindent \textbf{Colorization Diversity.} Since the color scribbles can be diverse, scribble-based video colorization should be a multimodal application. We illustrate some results from different color scribbles in Figure \ref{diversity}. It is clear that SVCNet can generate diverse colors from different input color scribbles.

\begin{table*}[t]
\begin{center}
\caption{SVCNet ablation study on DAVIS and Videvo datasets. The \textcolor{red}{\textbf{red}} color represents the best performance.}
\label{table_ablation}
\begin{tabular}{lccccccc}
\hline
\multirow{2}{*}{Ablation study setting} & \multirow{2}{*}{Compared item} & \multirow{2}{*}{Color scribble} & \multicolumn{3}{c}{DAVIS} & \multicolumn{2}{c}{Videvo} \cr & & & PSNR $\uparrow$ & SSIM $\uparrow$ & mIoU $\uparrow$ & PSNR $\uparrow$ & SSIM $\uparrow$ \\
\hline
1) w/o segmentation loss & Training scheme & \checkmark & 24.09 & 0.9378 & 0.8780 & 25.04 & 0.9464 \\
2) w/o CPNet pre-training & Training scheme & \checkmark & 19.19 & 0.8913 & 0.6464 & 20.27 & 0.9067 \\
3) w/o short-range connections & Temporal aggregation & \checkmark & 24.60 & 0.9495 & 0.8949 & 25.04 & 0.9535 \\
4) w/o the long-range connection & Temporal aggregation & \checkmark & 24.23 & 0.9479 & 0.8919 & 24.51 & 0.9511 \\
5) w/o short- and long-range connections & Temporal aggregation & \checkmark & 24.01 & 0.9481 & 0.8934 & 24.37 & 0.9498 \\
6) with 64$\times$128 resolution & Resolution & \checkmark & 24.47 & 0.9522 & 0.8768 & 25.26 & 0.9576 \\
7) with 128$\times$224 resolution & Resolution & \checkmark & 24.59 & 0.9533 & 0.8960 & 25.64 & 0.9589 \\
\hline
SVCNet & Full SVCNet & \checkmark & \textcolor{red}{25.71} & \textcolor{red}{0.9565} & \textcolor{red}{0.9104} & \textcolor{red}{26.30} & \textcolor{red}{0.9615} \\
\hline
\end{tabular}
\end{center}
\end{table*}


\begin{figure}[t]
\begin{center}
\centering
\includegraphics[width=0.95\linewidth]{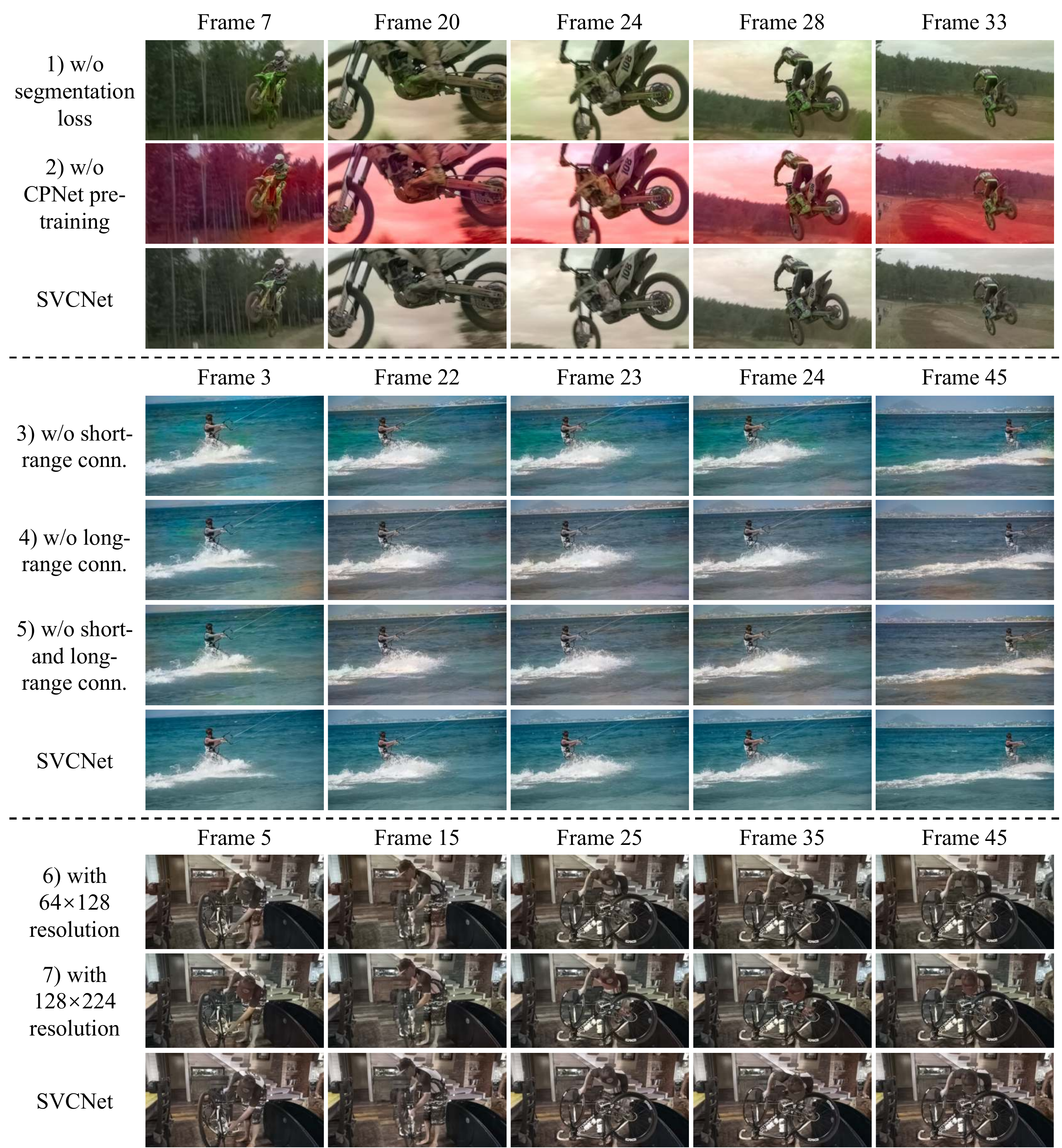}
\caption{Illustration of different ablation study settings. Samples are selected from the ``motocross-jump'', ``kite-surf'', and ``bike-packing'' samples of the DAVIS dataset, respectively.}
\label{ablation}
\end{center}
\end{figure}

\subsection{Ablation Study}

\noindent \textbf{Experiment Setting.} In order to demonstrate the effectiveness of components of the SVCNet, we define seven ablation study settings. The training hyper-parameters are unchanged excluding the ablation study items for each setting. The evaluation is on the original resolution on DAVIS and Videvo datasets, where all the settings share the same color scribbles. The details are concluded as follows:

1) w/o segmentation loss: Drop the segmentation loss $L_s^{CP}$;

2) w/o CPNet pre-training: Drop the CPNet warm-up pre-training stage on video datasets (i.e., only on the ImageNet);

3) w/o short-range connections: Drop the Refinement Module of SSNet with all short-range connections (i.e., $y_{t-3}^{ab}$, $y_{t-2}^{ab}$, $y_{t-1}^{ab}$, $y_{t+1}^{ab}$, $y_{t+2}^{ab}$, $y_{t+3}^{ab}$, and $z_{t-1}^{ab}$);

4) w/o the long-range connection: Drop the Correspondence Module with the long-range connection (i.e., $y_{1}^{ab}$);

5) w/o short- and long-range connections: Drop both short-range connections and the long-range connection;

6) with 64$\times$128 resolution: Change the input image resolution of SVCNet to 64$\times$128. The upsampling ratio of the Super-resolution Module is changed to 8;

7) with 128$\times$224 resolution: Change the input image resolution of SVCNet to 128$\times$224. The upsampling ratio of the Super-resolution Module is changed to 4.

The quantitative results are concluded in Table \ref{table_ablation} and some samples are illustrated in Figure \ref{ablation}.

\noindent \textbf{Training Scheme.} For settings 1) and 2), we exclude some parts in the training stages. If dropping the segmentation loss, there are color bleeding artifacts (e.g., some regions of the sky are colorized in green), as shown in Figure \ref{ablation}. If dropping the CPNet warm-up pre-training stage on video datasets, the colorization is wrong. Training all modules without warm-up pre-training is extremely difficult. It is because the CPNet cannot well colorize video frames that have different resolutions with images; meanwhile, the SSNet becomes ineffective when the given colorized frames from the CPNet are not good. In addition, as concluded in Table \ref{table_ablation}, the mIoU of setting 1) and results of all metrics of setting 2) decrease obviously.

\noindent \textbf{Temporal Aggregation.} For settings 3-5), we do not use some temporal information. In Figure \ref{ablation}, the colors of frames 22-24 results of setting 3) are not consistent since short-range connections are not used. The colors of far frame results of settings 4) and 5) are also not consistent with the frame 3 result. Since the long-range connection is not used and the color scribbles easily vanish when there are large motions, the colors for far frames are distorted. Also in Table \ref{table_ablation}, excluding temporal information leads to a giant decrease in both PSNR and SSIM values. The experiment results demonstrate that both short-range connections and the long-range connection are significant for the SVCNet.

\noindent \textbf{Resolution.} For settings 6) and 7), we change the image resolution. Though reducing the resolution can accelerate the inference speed, the performance of SVCNet drops obviously. Also in most cases, colorization applications do not need a very quick inference time. Considering the balance of colorization quality, inference speed, and memory cost, we use 256$\times$448 as the running resolution.

In conclusion, all the proposed training schemes, temporal aggregation, and network architecture are significant.

\begin{table}[t]
\begin{center}
\caption{Comparison of the number of parameters ($N_{param}$) and multiply accumulates (MACs) on a patch with a resolution of 256$\times$448 or 1024$\times$1792 ($MACs_{256}$ and $MACs_{1024}$) for SVCNet.}
\label{table_net}
\begin{tabular}{l|c|cc}
\hline
Module & $N_{param}$ & $MACs_{256}$ & $MACs_{1024}$ \\
\hline
CPNet & 91.475M & 251.690G & 251.690G \\
\hline
Correspondence Module & 26.942M & 162.102G & 162.102G \\
Refinement Module & 346.608K & 28.664G & 28.664G \\
Combination Module & 7.084M & 20.479G & 20.479G \\
Super-resolution Module & 306.432K & 33.096G & 55.292G \\
SSNet & 34.678M & 244.340G & 266.537G \\
\hline
SVCNet & 126.153M & 496.030G & 518.226G \\
\hline
\end{tabular}
\end{center}
\end{table}

\begin{figure*}[t]
\begin{center}
\centering
\includegraphics[width=0.95\linewidth]{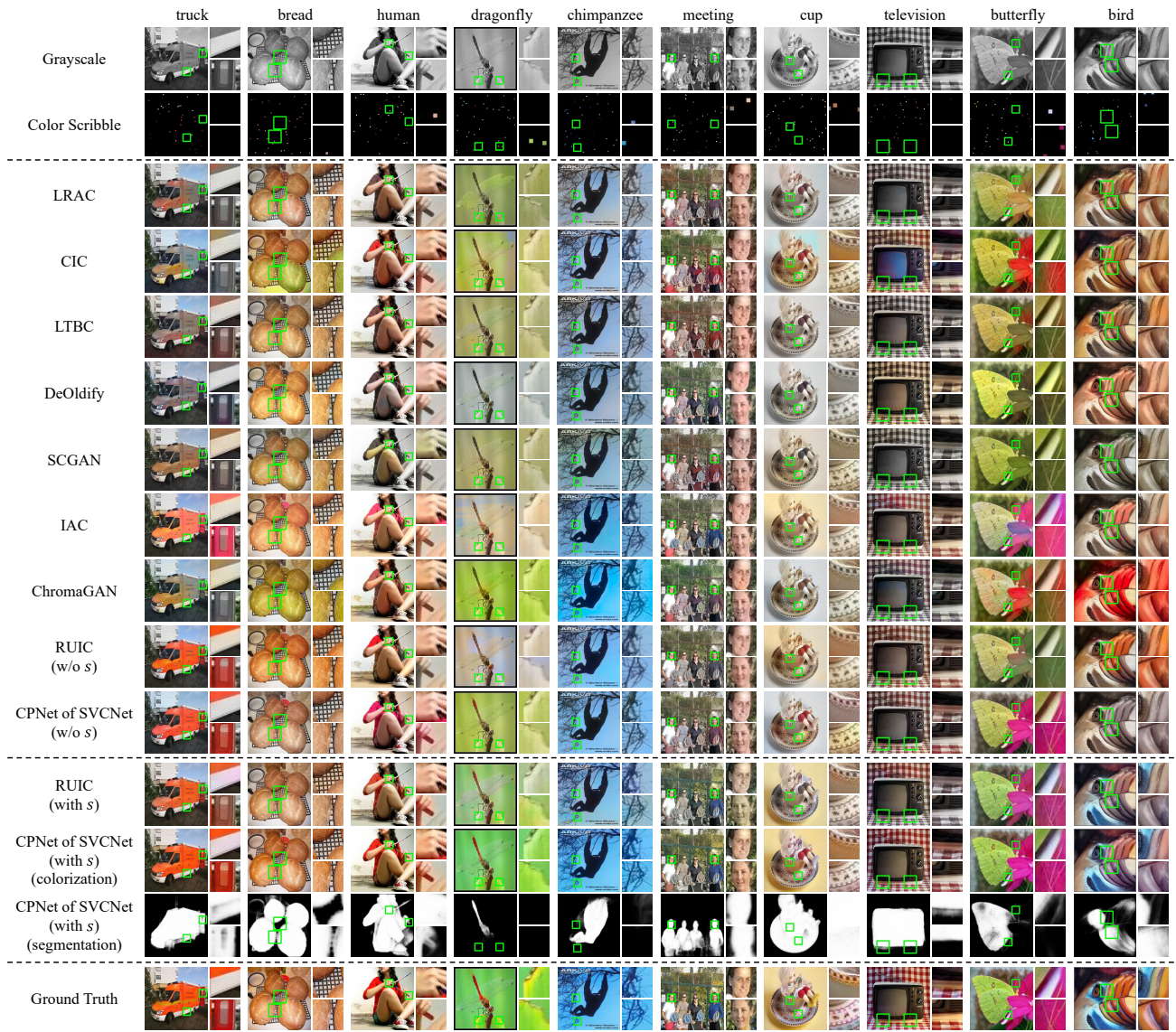}
\caption{Illustration of image colorization results of SVCNet and state-of-the-art methods, where RUIC \cite{zhang2017real} and SVCNet are scribble-based methods. The first and the second rows denote the grayscale and color scribbles, respectively. The last and the second last rows are the ground truth and predicted segmentation maps by SVCNet, respectively. The other rows include the colorization results of different methods. The patches are shown alongside full-resolution images.}
\label{image}
\end{center}
\end{figure*}

\begin{table}[t]
\begin{center}
\caption{Comparison of image colorization methods and the proposed CPNet of SVCNet on the ImageNet dataset.}
\label{table_image}
\begin{tabular}{l|c|cc}
\hline
Method & Color scribble & PSNR & SSIM \\
\hline
CIC \cite{zhang2016colorful} & / & 22.62 & 0.9153 \\
LTBC \cite{iizuka2016let} & / & 24.96 & 0.9464 \\
LRAC \cite{larsson2016learning} & / & 24.49 & 0.9229 \\
Pix2Pix \cite{isola2017image} & / & 23.39 & 0.9386 \\
DeOldify \cite{DeOldify} & / & 23.14 & 0.9194 \\
FAVC \cite{lei2019fully} & / & 22.96 & 0.9146 \\
ChromaGAN \cite{vitoria2020chromagan} & / & 23.67 & 0.9273 \\
SCGAN \cite{zhao2020scgan} & / & 23.93 & 0.9470 \\
IAC \cite{su2020instance} & / & 24.91 & 0.9110 \\
VCGAN \cite{zhao2022vcgan} & / & 24.58 & 0.9427 \\
RUIC \cite{zhang2017real} (w/o $s$) & / & 25.69 & 0.9526 \\
CPNet of SVCNet (w/o $s$) & / & \textcolor{green}{25.74} & \textcolor{green}{0.9577} \\
\hline
RUIC \cite{zhang2017real} (with $s$) & \checkmark & \textcolor{blue}{28.94} & \textcolor{blue}{0.9640} \\
CPNet of SVCNet (with $s$) & \checkmark & \textcolor{red}{31.40} & \textcolor{red}{0.9760} \\
\hline
\end{tabular}
\end{center}
\end{table}

\subsection{Computational Costs}

The computational costs for the SVCNet are concluded in Table \ref{table_net}. Based on the sparsity of color components, the majority of operations are on a fixed small resolution (i.e., 256$\times$448). Therefore, the computational costs for all modules except for the Super-resolution Module remain the same for different input resolutions. When changing the input resolution to 1024$\times$1792 from 256$\times$448, multiply accumulates (MACs) only increase by 22.197G, which is only 4.5\% of MACs of 256$\times$448 resolution (496.030G). This design makes the SVCNet memory-friendly. In addition, using small architectures for the Refinement Module and the Super-resolution Module is enough. The sum of their parameters accounts for less than 1\% of the overall parameters of the whole SVCNet. Please refer to Figure \ref{net} (b) and the base channels for them are 16 and 32, respectively. Though small architectures are used, they fulfill the targets well, as shown in Figure \ref{net} (c) and (d), respectively.

\begin{figure*}[t]
\begin{center}
\centering
\includegraphics[width=0.95\linewidth]{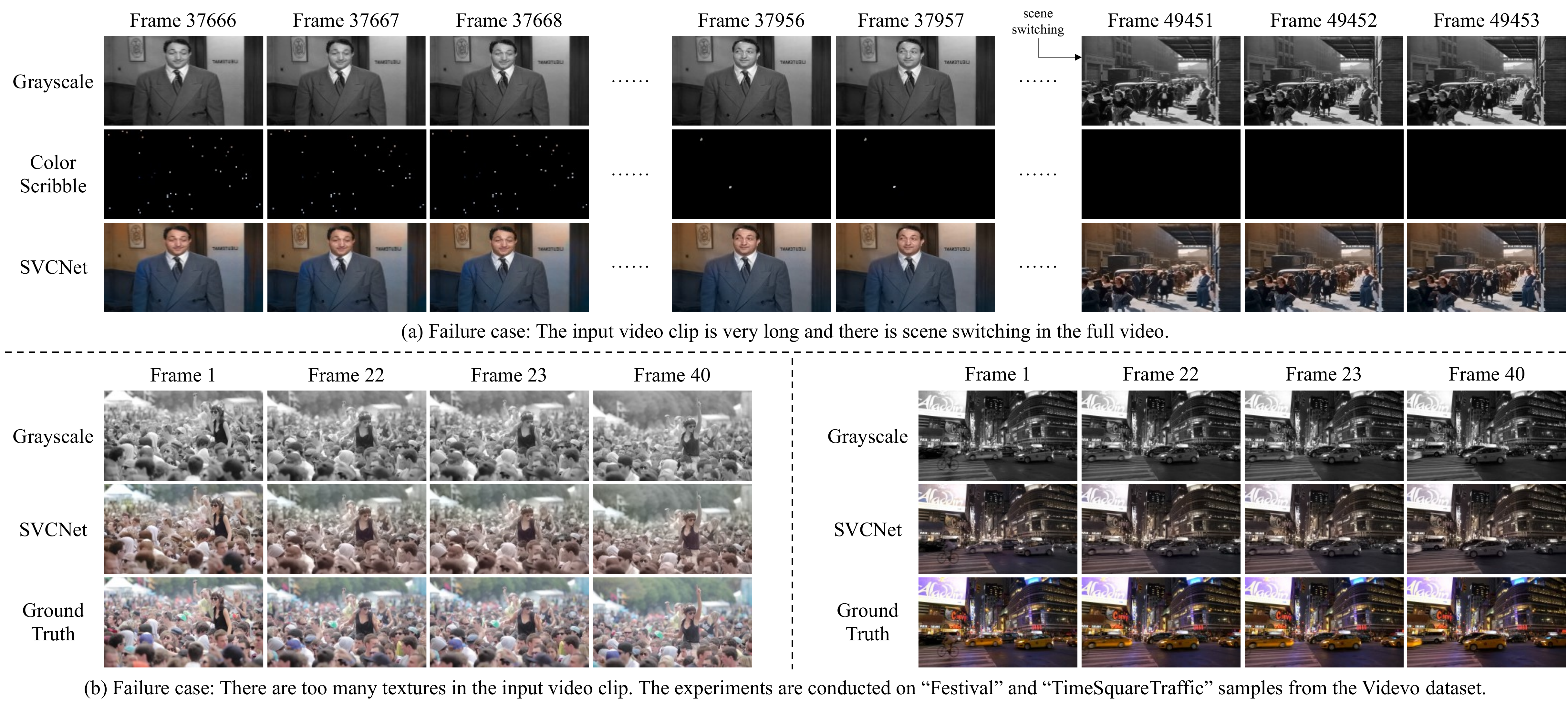}
\caption{Illustration of two common failure cases: (a) The input video clip is too long and there is scene switching; (b) There are too many textures. The frames in sub-figure (a) are from a 1948 grayscale film ``The Naked City'' with an FPS of 24. The input color scribbles are omitted for sub-figure (b).}
\label{failure}
\end{center}
\end{figure*}

\subsection{Image Colorization Experiments}

\noindent \textbf{Experiment Setting.} In order to further demonstrate the colorization quality of SVCNet, we compare the CPNet of SVCNet with the following baselines:

1) Fully-automatic methods: CIC \cite{zhang2016colorful}, LTBC \cite{iizuka2016let}, LRAC \cite{larsson2016learning}, Pix2Pix \cite{isola2017image}, DeOldify \cite{DeOldify}, FAVC \cite{lei2019fully}, ChromaGAN \cite{vitoria2020chromagan}, SCGAN \cite{zhao2020scgan}, IAC \cite{su2020instance};

2) Scribble-based method: RUIC \cite{zhang2017real}.

All the methods are trained on ImageNet 1.3 million training set and evaluated on ImageNet 10000 validation set, as defined by \cite{zhang2016colorful, larsson2016learning, zhao2020scgan}. All the methods are trained and evaluated on 256$\times$256 image resolution. We use 40 color scribbles for SVCNet and RUIC in the experiment.

\noindent \textbf{Qualitative Analysis.} The qualitative samples are illustrated in Figure \ref{image}. Firstly, compared with other methods, there are fewer color bleeding artifacts in the CPNet results (i.e., the color of an object does not permeate through other objects). It is because we use a segmentation branch and a segmentation loss, which helps the network focus on the key objects and separate them from other objects. In the contrast, there are obvious artifacts for the other methods, e.g., ``truck'' patches from LRAC, CIC, LTBC, and ``chimpanzee'', ``meeting'' patches from RUIC, etc. Secondly, there is fewer color confusion issue (i.e., the colors are semantically wrong for some objects) in the CPNet results even when there are no given color scribbles. For instance, the background of the ``dragonfly'' sample is colorized to blue for IAC, DeOldify, and RUIC (w/o $s$). The background of the ``cup'' sample is colorized not consistently for CIC, IAC, and RUIC (w/o $s$). The colors of ``truck'' from LRAC, CIC, LTBC, and ChromaGAN are not consistent. However, the colors of these samples are more reasonable for the CPNet. In conclusion, the proposed CPNet has a stronger ability to perform scribble-based image colorization, which serves as a powerful backbone for the SVCNet.

\noindent \textbf{Quantitative Analysis.} The quantitative results are concluded in Table \ref{table_image}. On one hand, CPNet obtains better performance than the state-of-the-art scribble-based image colorization method RUIC either without color scribbles (w/o $s$) or with color scribbles (with $s$). Especially when using color scribbles, CPNet largely outperforms RUIC, e.g., 2.46 higher PSNR and 0.0120 higher SSIM. Based on the powerful CPNet, SVCNet has the potential to colorize videos with high quality. On the other hand, CPNet outperforms existing methods when using color scribbles. It is because CPNet can well utilize the information from the color scribbles to guide the colorization. Compared with conventional architectures, SVCNet directly uses a pre-trained semantic feature encoder to extract features and has a decoder segmentation branch to predict the segmentation map, which helps the network address the color bleeding artifacts. The network designs also contribute to better performance.

\subsection{Failure Cases and Discussion}

The SVCNet can produce high-quality colorful videos by propagating the color scribbles to the grayscale videos. The results are often not plausible when: 1) the video clip is very long and there is scene switching in the full video; 2) there are too many textures, as illustrated in Figure \ref{failure}.

Firstly, we use a real legacy video for experiments, as shown in Figure \ref{failure} (a). When we give the color scribbles for the first frame of a scene (e.g., Frame 37666), the color styles of results for far frames in the scene (e.g., Frames 37666 and 37957) are consistent with the first frame. However, when we colorize the far frames (e.g., Frame 49451, which is approximately 8 minutes later than Frame 37666), the details are not good. The color style of Frame 49451 is also very similar to the first colorized frame (i.e., Frame 37666) because the SVCNet uses only the long-range connection for reference; however, the styles of the two distinct scenes are not always similar. We assume that new color scribbles should be given for far keyframes with a keyframe detection algorithm \cite{yan2018deep}. Secondly, when there are a lot of details in grayscale videos, the SVCNet is hard to colorize the tiny objects, as shown in Figure \ref{failure} (b). In the future, we will develop more powerful color propagation techniques and make SVCNet robust to complicated textures.

\section{Conclusion}

In this paper, we present the first scribble-based video colorization framework called SVCNet. It includes two sequential sub-networks called color propagation network (CPNet) and spatiotemporal smoothing network (SSNet). The CPNet performs accurate image colorization based on the given color scribbles. Utilizing two feature encoders, it effectively extracts semantics and fuses the information of color scribbles and grayscale images. It also contains two decoder branches, where one for producing color embeddings and the other for generating corresponding segmentation maps. By enforcing the multi-task losses at training, the segmentation branch helps CPNet alleviate color bleeding artifacts. The SSNet post-processes the output from the CPNet, which aggregates short-range connections (neighboring colorized frames), the long-range connection (the first colorized frame), and information of the current time step (the current CPNet output) to achieve good temporal consistency. In addition, we notice that color embeddings are sparse so we set the majority of operations to a fixed small resolution and we use a Super-resolution Module to recover the larger resolution for HD video applications at the tail of SVCNet. Finally, we compare SVCNet with several state-of-the-art image colorization and video colorization methods. The results demonstrate that SVCNet produces more realistic results and encounters fewer color bleeding artifacts than existing methods.


%

%



\ifCLASSOPTIONcaptionsoff
  \newpage
\fi

{
\bibliographystyle{IEEEtran}
\bibliography{IEEEabrv,mybibfile}
}

\vspace{-8mm}

\begin{IEEEbiography}[{\includegraphics[width=1in,height=1.5in,clip,keepaspectratio]{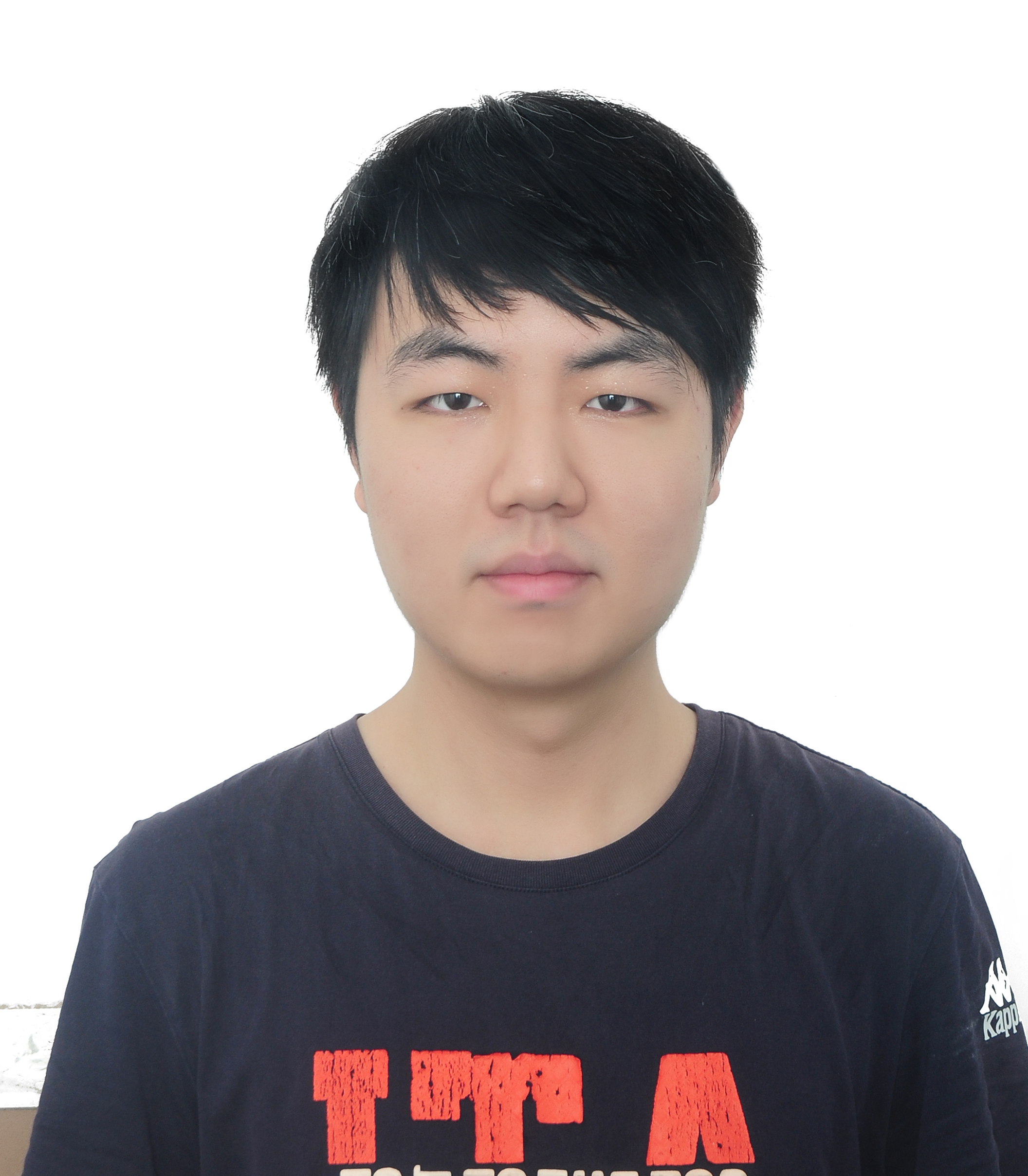}}]{Yuzhi Zhao}

(S’19) received the B.Eng. degree in Electronic and Information Engineering from Huazhong University of Science and Technology (HUST), 2018 and the Ph.D. degree in Electronic Engineering from City University of Hong Kong (CityU) in 2023. His research interests include low-level vision, computational photography, generative models, and representation learning. He serves as a peer-reviewer in international conferences and journals such as CVPR, ICCV, ECCV, WACV, ACCV, ICASSP, ICIP, and several IEEE Transactions.

\end{IEEEbiography}

\vspace{-8mm}

\begin{IEEEbiography}[{\includegraphics[width=1in,height=1.25in,clip,keepaspectratio]{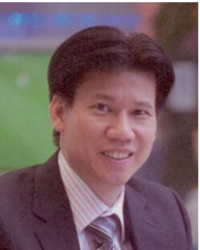}}]{Lai-Man Po}

(M’92–SM’09) received the B.S. and Ph.D. degrees in electronic engineering from the City University of Hong Kong, Hong Kong, in 1988 and 1991, respectively. He has been with the Department of Electronic Engineering, City University of Hong Kong, since 1991, where he is currently an Associate Professor of Department of Electrical Engineering. He has authored over 150 technical journal and conference papers. His research interests include image and video coding with an emphasis deep learning based computer vision algorithms.

Dr. Po is a member of the Technical Committee on Multimedia Systems and Applications and the IEEE Circuits and Systems Society. He was the Chairman of the IEEE Signal Processing Hong Kong Chapter in 2012 and 2013. He was an Associate Editor of HKIE Transactions in 2011 to 2013. He also served on the Organizing Committee, of the IEEE International Conference on Acoustics, Speech and Signal Processing in 2003, and the IEEE International Conference on Image Processing in 2010.

\end{IEEEbiography}

\vspace{-8mm}

\begin{IEEEbiography}[{\includegraphics[width=1in,height=1.25in,keepaspectratio]{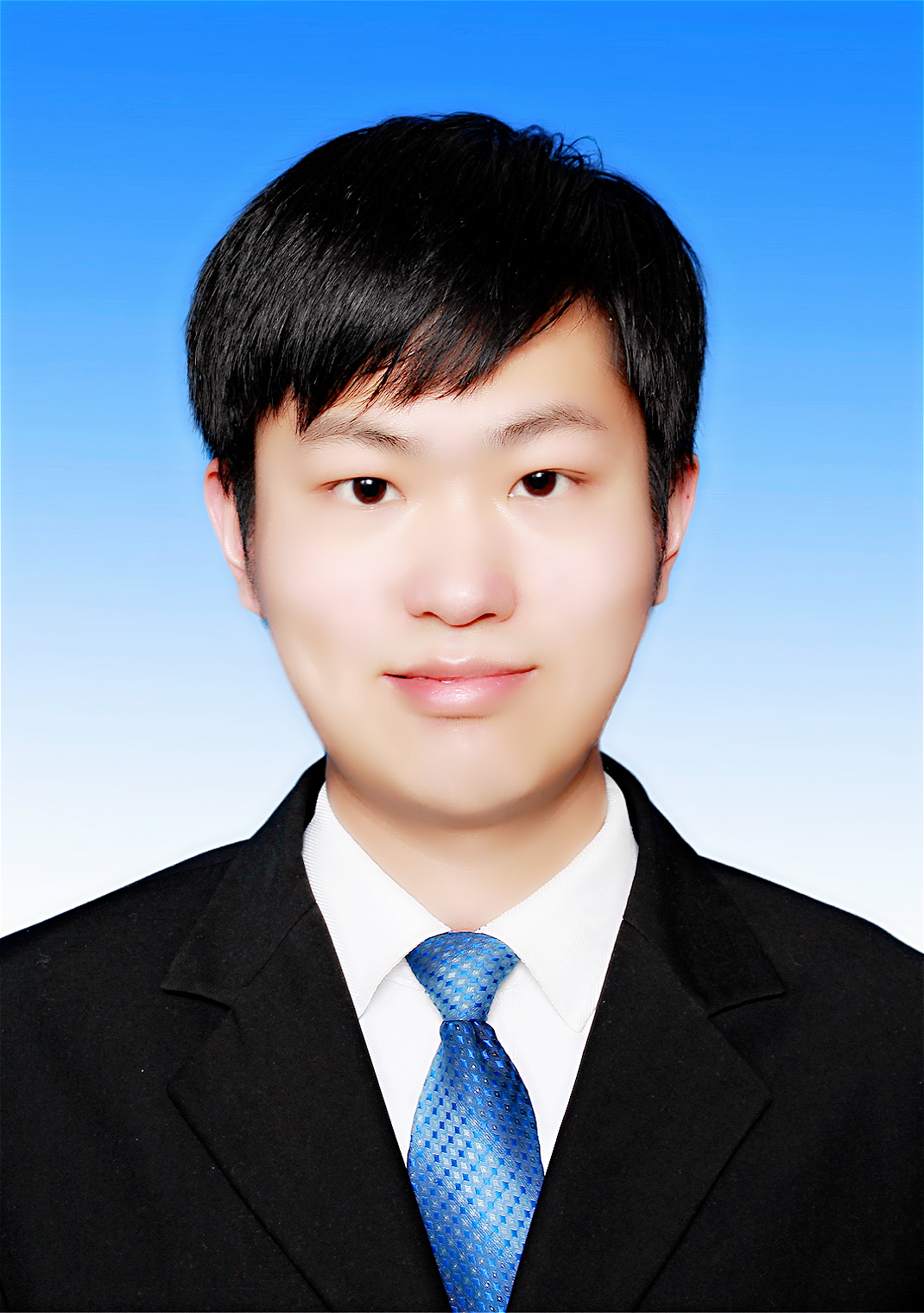}}]{Kangcheng Liu}

(M’19) is currently a Research Fellow at Nanyang Technological University (NTU), Singapore. He received the B.Eng. degree from the Harbin Institute of Technology (HIT) in 2018 and the Ph.D. degree in Mechanical and Automation Engineering from The Chinese University of Hong Kong (CUHK) in 2022 focusing on robotics. He served as a program committee member of IEEE ICRA and the reviewer of CVPR, ICCV, ECCV, ACM SIGGRAPH,  IJCV, T-RO, and T-PAMI. 
 
\end{IEEEbiography}

\vspace{-5mm}

\begin{IEEEbiography}[{\includegraphics[width=1in,height=1.25in,clip,keepaspectratio]{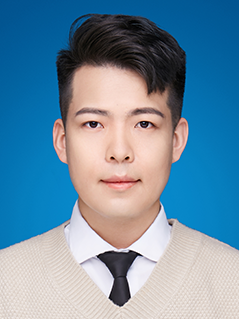}}]{Xuehui Wang}

(S’21) received the B.S. Degree from Shandong University, China, and the Master degree in the School of Computer Science from Sun Yat-sen Uninversity, China, in 2018 and 2021, respectively. He is currently pursuing the Ph.D. degree at Artificial Intelligence Institute, Shanghai Jiao Tong University, China. His research interests include computer vision (super resolution, instance segmentation), deep learning.

\end{IEEEbiography}

\vspace{-5mm}

\begin{IEEEbiography}[{\includegraphics[width=1in,height=1.25in,clip,keepaspectratio]{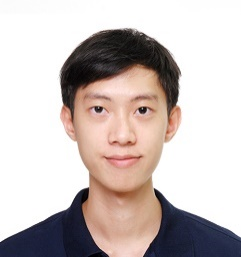}}]{Wing-Yin Yu}

(S’21) received the B.Eng. degree in Information Engineering from City University of Hong Kong, in 2019. He is currently pursuing the Ph.D. degree at Department of Electronic Engineering at City University of Hong Kong. His research interests are deep learning and computer vision.

\end{IEEEbiography}

\vspace{-5mm}

\begin{IEEEbiography}[{\includegraphics[width=1in,height=1.25in,clip,keepaspectratio]{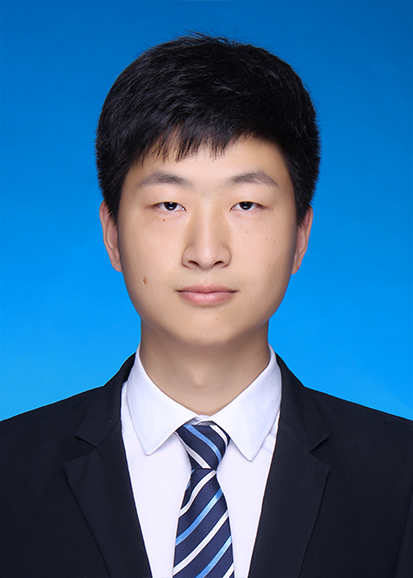}}]{Pengfei Xian}

(S’21) received the B.Eng. degree in electrical engineering from Harbin Institute of Technology, Harbin, China, in 2017. He is currently pursuing the Ph.D. degree in electrical engineering at City University of Hong Kong. His research interest includes instance and sematic segmentation on images and videos, together with reinforcement learning applications.

\end{IEEEbiography}

\vspace{-5mm}

\begin{IEEEbiography}[{\includegraphics[width=1in,height=1.7in,clip,keepaspectratio]{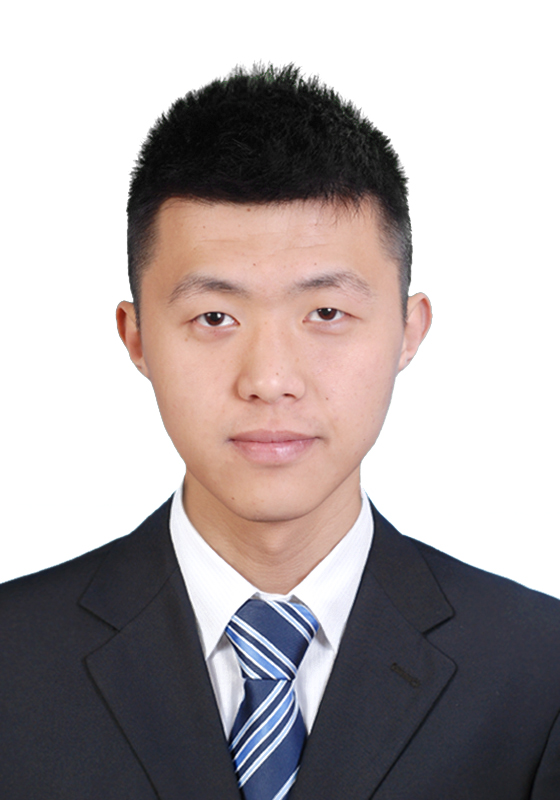}}]{Yujia Zhang}

received the B.Eng. degree in electrical engineering and automation in Huazhong University of Science and Technology in 2015, the M.Sc. degree in electrical engineering in South China University of Technology in 2018, and the Ph.D. degree in electrical engineering in City University of Hong Kong in 2022. He is currently an engineer with the Tencent Video, Tencent Holdings Ltd. His current research interests include computer vision and video understanding.

\end{IEEEbiography}

\vspace{-5mm}

\begin{IEEEbiography}[{\includegraphics[width=1in,height=1.7in,clip,keepaspectratio]{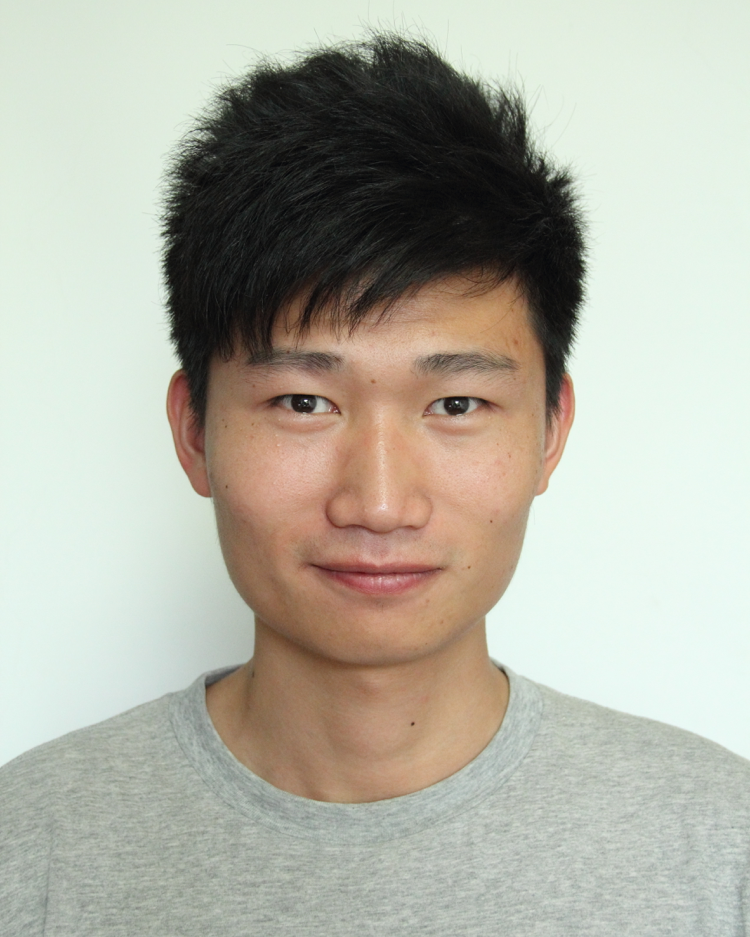}}]{Mengyang Liu}

received the B.Eng. degree in optoelectronic engineering from the Shanghai University of Electric Power, Shanghai, China, in 2014, and the M.Sc. degree in electronic and information engineering and the Ph.D. degree from the City University of Hong Kong, in 2015 and 2019, respectively. He is currently an engineer with the Tencent Video, Tencent Holdings Ltd. His research interests include image and video processing, video embedding and retrieval, computer vision, and machine learning.

\end{IEEEbiography}

%




\end{document}